\documentclass[11pt,twoside]{article}
\usepackage{amsmath, amssymb, amscd, amsthm, amsfonts}
\usepackage{hyperref}
\usepackage{braket}
\usepackage[ruled,linesnumbered]{algorithm2e}
\usepackage{color}
\usepackage{geometry}
\usepackage{authblk}
\usepackage{xcolor}
\hypersetup{
    colorlinks,
    linkcolor={red!50!black},
    citecolor={blue!50!black},
    urlcolor={blue!80!black}
}

\newtheorem{theorem}{Theorem}
\newtheorem{example}{Example}
\newtheorem{assumption}{Assumption}

\newtheorem{lemma}{Lemma}
\newtheorem{corollary}{Corollary}
\newtheorem{definition}{Definition}
\newtheorem*{remark}{Remark}

\newenvironment{Mechanism}[1][htb]
{
    \begin{algorithm}[#1]}{\end{algorithm}}

\newcommand{\bb}{\mathbb}
\newcommand{\proj}{\Pi_{\bar{\mathcal{M}}}}

\newcommand \dd[1]{ \,\textrm d{#1}}
\newcommand\numberthis{\addtocounter{equation}{1}\tag{\theequation}}
\makeatletter
\newcommand\email[2][]%
{\newaffiltrue\let\AB@blk@and\AB@pand
    \if\relax#1\relax\def\AB@note{\AB@thenote}\else\def\AB@note{\relax}%
    \setcounter{Maxaffil}{0}\fi
    \begingroup
    \let\protect\@unexpandable@protect
    \def\thanks{\protect\thanks}\def\footnote{\protect\footnote}%
    \@temptokena=\expandafter{\AB@authors}%
    {\def\\{\protect\\\protect\Affilfont}\xdef\AB@temp{#2}}%
    \xdef\AB@authors{\the\@temptokena\AB@las\AB@au@str
        \protect\\[\affilsep]\protect\Affilfont\AB@temp}%
    \gdef\AB@las{}\gdef\AB@au@str{}%
    {\def\\{, \ignorespaces}\xdef\AB@temp{#2}}%
    \@temptokena=\expandafter{\AB@affillist}%
    \xdef\AB@affillist{\the\@temptokena \AB@affilsep
        \AB@affilnote{}\protect\Affilfont\AB@temp}%
    \endgroup
    \let\AB@affilsep\AB@affilsepx
}
\makeatother
\title{Truthful Generalized Linear Models}

\author[1]{Yuan Qiu\footnote{The paper was done when Yuan Qiu was a research intern at King Abdullah University of Science and Technology.}}
\author[2]{Jinyan Liu}
\author[3,4,5]{Di Wang}
\affil[1]{College of Computing, Georgia Institute of Technology}
\affil[2]{School of Computer Science and Technology, Beijing Institute of Technology}
\affil[3]{Division of CEMSE, King Abdullah University of Science and Technology}
\affil[4]{SDAIA-KAUST Center of Excellence in Data Science and Artificial Intelligence} 
\affil[5] {Computational Bioscience Research Center}
\email{yuan.qiu@gatech.edu, jyliu@bit.edu.cn, di.wang@kaust.edu.sa}
\date{}

\begin{document}
    \maketitle
    \begin{abstract}
        In this paper we study estimating Generalized Linear Models (GLMs) in the case where the agents (individuals) are strategic or self-interested and they concern about their privacy when reporting data. Compared with the classical setting, here we aim to  design mechanisms that can both incentivize most agents to truthfully report their data and preserve the privacy of individuals' reports, while their outputs should also close to the underlying parameter. In the first part of the paper, we consider the case where the covariates are sub-Gaussian and the responses are heavy-tailed where they only have the finite fourth moments. First, motivated by the stationary condition of the 
        maximizer of the likelihood function, we derive a novel private and closed form estimator. Based on the estimator, we propose a mechanism which has the following properties via some appropriate design of the computation and payment scheme for several canonical models such as linear regression, logistic regression and Poisson regression:  (1) the mechanism is $o(1)$-jointly differentially private (with probability at least $1-o(1)$); (2) it is an $o(\frac{1}{n})$-approximate Bayes Nash equilibrium for a $(1-o(1))$-fraction of agents to truthfully report their data, where $n$ is the number of agents; (3) the output could achieve an error of $o(1)$ to the underlying parameter; (4) it is individually rational for a $(1-o(1))$ fraction of agents in the mechanism ; (5) the payment budget required from the analyst to run the mechanism
        is $o(1)$. In the second part, we consider the linear regression model under more general setting where both covariates and responses are heavy-tailed and only have finite fourth moments. By using an $\ell_4$-norm shrinkage operator, we propose a private estimator and payment scheme which have similar properties as in the sub-Gaussian case.  
    \end{abstract}
    
    \section{Introduction}
    \label{sec:introduction}
    As one of the most fundamental models in statistics and machine learning, Generalized Linear Models (GLMs) have been intensively studied and widely applied to many areas such as medical trails~\cite{schwemer2000general}, census surveys~\cite{nordberg1989generalized} and crowdsourcing~\cite{alvaro2015crowdsourcing}. Among these studies, it is always assumed that the analysts hold high-quality data, which is essential to the success of GLMs. However, in many scenarios, such as medical trails and census surveys,  data of interest may contain sensitive information and thus they may be collected from strategic and self-interested individuals who are concerned with their privacy. In this case, data providers (agents)\footnote{In this paper, individuals, data providers and participants are the same and all represent agents. } may be unwilling to truthfully report their data, which will result in the failure of estimating the underlying model. Thus, compared with the classical statistical setting,  it is necessary to model utility functions of individuals and to design mechanisms that can output accurate estimators, preserve the privacy of individuals reports, and provide proper incentives to encourage most individuals to truthfully report their data to the analyst.

    In general, the goal of solving the problem can be divided into two interconnected components -- data acquisition and privacy-preserving data analysis. On one hand, an analyst will pay individuals (agents) in compensation for possible privacy violation. He/She should pay each agent strategically according to how well the reported data aligned with the underlying statistical model and peers' data, but meanwhile he/she needs to minimize the total payment budget. On the other hand, the analyst needs to perform privacy-preserving computation on the reported data to learn the underlying model accurately. Thus, there is a tradeoff between the accuracy of the estimator and the amount of payment budget required to compensate participants. In this paper, we provide the first study on this tradeoff for GLMs by proposing several Differentially Private (DP) mechanisms  under different settings. Specifically, our contribution can be summarized as follows. 
    \begin{itemize}
        \item In the first part of the paper, we focus on GLMs where the distributions of covariates are sub-Gaussian and the  distributions of response are heavy-tailed (only have finite fourth moments). First, based on stationary condition of the maximizer  of the likelihood function for GLMs, we derive a closed form estimator and privatize the estimator to make it satisfies DP (with high probability). Based on the DP estimator, we propose a general design of computation and payment scheme. Specifically, for some canonical models such as linear regression, logistic regression and Poisson regression, our mechanism has the following properties (if we assume that the dimension of the data is $O(1)$): 
        \begin{enumerate}
            \item The mechanism preserves privacy for individuals' reported data, i.e., the output of the mechanism is $o(1)$-Jointly Differentially Private (Definition \ref{def:rjdp}) with probability $1-O(n^{-{\Omega(1)}})$, where $n$ is the number of participants (agents). 
            \item The private estimator of the mechanism is $o(1)$-accurate, i.e., when the number of agents increases, our private estimator will be sufficiently close to the underlying parameter. 
            \item The mechanism is asymptotically truthful, i.e., it is an $o(\frac{1}{n})$-approximate Bayes Nash equilibrium for a $(1-o(1))$-fraction of agents to truthfully report their data. 
            \item The mechanism is asymptotically individually rational, i.e., the utilities of a $(1-o(1))$-fraction of agents are non-negative. 
            \item The mechanism only requires $o(1)$ payment budget, i.e.,  when the number of participants increases, the total payment tends to zero. 
        \end{enumerate}
        \item One disadvantage of the previous method is that the it relies on the assumption that the distributions of covariates are sub-Gaussian, which may not hold in some scenarios. To address this issue, in the second part we consider a more general setting where the distributions of both covariates and responses are heavy-tailed. Specifically, we focus on the linear regression model and provide a private estimator by applying an $\ell_4$-norm shrinkage operator to each covariate. Based on the private estimator and the idea of the above method, we present a mechanism which has similar properties as in the sub-Gaussian data case. 
    \end{itemize}
    Due to space limit, all the proofs and technical lemmas are included in the Appendix. 

   %\vspace{-0.2in}
    \section{Related work}
    %\vspace{-0.1in}
    \label{subsec:related_work}
    
    Start from~\cite{ghosh2011selling}, there is a long list of work studies data acquisition from agents that have privacy concern from different perspectives~\cite{ligett2012take,fleischer2012approximately,nissim2012privacy,ghosh2014buying,fallah2022optimal}. However, most of them do not consider statistical estimation problems.~\cite{cummings2015truthful} is the work that is most closest to ours. It focuses on estimating the linear regression model from self-interested agents that have privacy concern. However, there are several critical differences compared with our work. First, the method in~\cite{cummings2015truthful} is based on the optimal solution of linear regression, which has a closed form and thus cannot be extended to GLMs as the optimal solution of GLMs does not have closed form in general. Secondly,~\cite{cummings2015truthful} needs strong assumptions on the data distribution to achieve the privacy guarantee, i.e., it need to assume that the $\ell_2$-norm of the covariates and responses are bounded, while in this paper we extend the setting to the heavy-tailed case. Such extension is non-trivial as here we use the $\ell_4$-norm shrinkage (see Section~\ref{subsec:linear} for details) to preprocess the covariates. Recently~\cite{kong2020information} also considers mean estimation and linear regression estimation from agents with privacy concern. However, there is no DP guarantee for their methods. Thus, it is incomparable with our work.
    
    In the classical setting of estimating GLMs in the DP model, there are numerous approaches, such as~\cite{bassily2021differentially,jain2014near,song2021evading,chaudhuri2011differentially,cai2020cost}.  However, all of them are based on adding noise to the output of some optimization methods,  privatizing the objective function or adding noise to gradients in optimization methods, which cannot be adopted to our problem as data acquisition is just a single-round interactive procedure between analyst and  agents, while those above approaches need multiple rounds of interactions. To address the issue, we propose a novel and non-trivial private estimator for GLMs. Compared with the previous approaches, our estimator has a closed-form expression and can be gotten via single round of interaction. This is similar to the linear regression case and we believe that it can also be used in other related problems. 
    
    Besides the privacy concern, statistical estimation from strategic agents also has been studied in a variety of different contexts. For example,~\cite{chen2018strategyproof} studies linear regression in the case where agents may intentionally introduce errors to maximize their own benefits and presents several group strategyproof linear regression mechanisms, which are later extended to classification problems~\cite{chen2020learning}. ~\cite{hardt2016strategic} proposes learning classifiers that are robust to agents strategically misreporting their feature vectors to trick the algorithm into misclassifying them. In~\cite{cai2015optimum}, the authors study fitting linear regression model in the case where agents can only manipulate their costs instead of their data.

    \vspace{-0.2in}
    \section{Preliminaries}
      \vspace{-0.1in}
    \label{sec:preliminaries}
    {\bf Notations:} Given a matrix $X\in \bb{R}^{n\times d}$, we denote its $i$-th row by $\mathbf{x}_i^T$ and its $(i,j)$-th entry by $[X]_{ij}$. For a vector $v$, denote $[v]_j$ or $v_{ij}$ as its $j$-th coordinate. For any $p\in [1,\infty]$, let $\|X\|_p$ denote its $p$-norm, i.e., $\|X\|_p:=\sup_{y\neq 0}\frac{\|Xy\|_p}{\|y\|_p}$. For an event $A$,  we denote the indicator as $\mathbf{1}_A$ where  $\mathbf{1}_A=1$ if $A$ occurs, otherwise $\mathbf{1}_A=0$. The sign function of a real number $x$ is a piecewise function which is defined as $ \mathrm{sgn}(x)=-1$ if $x<0$; $\mathrm{sgn}(x)=1$ of $x>0$; and $\mathrm{sgn}(x)=0$ if $x=0$. 
      \vspace{-0.1in}
    \subsection{Problem Setting} 
    \label{subsec:problem_setting}
    
    Suppose that there is a data universe $\mathcal{D}=\mathcal{X}\times \mathcal{Y}\subseteq \mathbb{R}^d\times \mathbb{R}$ and  $n$ agents in the population. The $i$-th agent has a feature vector (covariate) $\mathbf{x}_i\in \mathcal{X} $, and a response variable $y_i\in \mathcal{Y}$. We assume $\{(\mathbf{x}_i, y_i)\}_{i=1}^n$ are i.i.d.
    sampled from a Generalized Linear Model (GLM). That is, $\mathbf{x}_i$ are i.i.d. random vectors drawn from some unknown distribution $\mathcal{F}$ and  there exists a $\theta^*\in \mathbb{R}^d$ such that the conditional probability function $\mathcal{G}_{\mathbf{x}_i}=p(\cdot|\mathbf{x}_i)$ of $y_i$ has the following parameterized form:
    \begin{align*}
        p(y_i|\mathbf{x}_i;\theta^*)
        =\exp\left\{\frac{y_i\langle\mathbf{x}_i,\theta^{*}\rangle-A(\langle\mathbf{x}_i,\theta^{*}\rangle)}{\phi}+c(y_i,\phi)\right\},
        \numberthis\label{def:glm_pdf}
    \end{align*}
    where $\phi\in\bb{R}$ is a fixed and known scale parameter, $c(\cdot, \cdot)$ is some known function and $A(\cdot)$ is the link function. %To make the parametrization to be valid, the conditional density should be normalizable, so that $A(\langle\mathbf{x}_i, \theta^{*}\rangle)<+\infty$.
    We assume that function  $A$ is twice differentiable, and its derivative function $A^{\prime}$ is monotonic increasing. It is well-known that $\bb{E}[y|\mathbf{x}_i, \theta^{*}]=A^{\prime}(\langle\mathbf{x}_i,\theta^{*}\rangle)$ and $\mathrm{var}[y|\mathbf{x}_i, \theta^{*}]=A^{\prime\prime}(\langle\mathbf{x}_i, \theta^{*}\rangle)\phi$ (where $A''(\cdot)$ is the second derivative function of $A(\cdot)$). Note that GLMs include several canonical models such as linear regression, logistic regression and Poisson regression (see Section~\ref{subsec:examples} for details). In this paper, we focus on the low dimensional case which means $n\gg d$ and we make the following assumptions on the parameter of interest.
    
    \begin{assumption}
        \label{assump1:theta&lowD}
        Throughout the paper, we assume that the model parameter $\theta^{*}\in\mathbb{R}^d$ is drawn from a prior distribution $p(\theta)$, and $\|\theta^{*}\|_2\leq\tau_{\theta}$ with some (known) constant $\tau_{\theta}>0$.
    \end{assumption}
    There is an analyst who aims to estimate the underlying parameter in~\eqref{def:glm_pdf} from agents' data. That is, she/he wants to estimate $\theta^*$ based on data $D=\{D_i=(\mathbf{x}_i, y_i)\}_{i=1}^n$. As we mentioned previously, here we consider the case that the agents are strategic or self-interested, and they concern on the privacy when reporting their data. Specifically, we assume that each agent is characterized by a privacy cost coefficient $c_i\in\bb{R}_{+}$.  Higher value of $c_i$ indicates that agent $i$ concerns more about the privacy violation due to truthfully reporting $y_i$ to the analyst. %Given $\mathbf{x}_i$, the corresponding $y_i$ is a random variable drawn from some unknown distribution $\mathcal{G}_{\mathbf{x}_i}$ that depends on $\mathbf{x}_i$.
    Thus, due to the privacy concern, each agent $i$ can manipulate his/her response $y_i$.~\footnote{However, we assume that each agent $i$ cannot manipulate her/his feature vector $\mathbf{x}_i$, which has the same setting as in~\cite{cummings2015truthful}.} If we denote $\hat{y}_i$ as the the reported response, $\hat{D}_i=(\mathbf{x}_i, \hat{y}_i)$ as the reported data, and $\sigma_i$ as the reporting strategy, i.e., $\hat{y}_i=\sigma_i(D_i)$. Then the main goal of the analyst is to estimate the  parameter vector $\theta^{*}\in \bb{R}^d$ based on the reported data $\hat{D}=\{\hat{D}_i\}_{i=1}^{n}$. Moreover, as agents may lie about their private responses $y_i$, the analyst need to construct a payment rule $\pi: \mathcal{D}^n\to \Pi^n$ that encourages truthful reporting, i.e., misreporting the response $y_i$ will lead to lower received payment $\pi_i$.
    
    Overall, the analyst aims to a design a truthful mechanism $\mathcal{M}$ which takes the reported data $\hat{D}$ as input, and outputs an estimator $\bar{\theta}$ of $\theta^*$ and a set of non-negative payments $\{\pi_i\}_{i=1}^{n}$ for each agent. To make the mechanism incentivize truthful participation of most agents,  there should be some privacy guarantees for the reports provided by agents.  Informally, we seek private mechanisms that allow accurate estimation of $\theta^*$ and require only asymptotically small payment budget. All the above build upon the agents’ rational behaviors and the privacy model, which will be discussed in details in the following sections.

    \subsection{Differential Privacy}
    \label{subsec:differential_privacy}
    In this section, we define the desired criteria of privacy protection. We adopt some relaxations of the canonical notion of {Differential Privacy} (DP).
    
    \begin{definition}[$\varepsilon$-Differential Privacy~\cite{dwork2006calibrating}]
        \label{def:dp}
        Given a data universe $\mathcal{D}$ and any positive integer $n$, we say that two $n$-size datasets $D,D'\subseteq \mathcal{D}^n$ are neighbors if they differ by only one data sample, which is denoted as $D \sim D'$. A randomized algorithm $\mathcal{A}$ is $\epsilon$-differentially private (DP) if for all neighboring datasets $D,D'$ and for all events $S$ in the output space of $\mathcal{A}$, we have \footnote{All the methods and results in this paper can be extended to $(\epsilon, \delta)$ version of DP by adding Gaussian noise. For simplicity, we omit them here. } $$\mathbb{P}(\mathcal{A}(D)\in S)\leq e^{\epsilon} \mathbb{P}(\mathcal{A}(D')\in S).$$
    \end{definition}
    The definition of DP guarantees that the distributions of $\mathcal{A}(D)$ and $\mathcal{A}(D^{\prime})$ are almost indistinguishable. In other words, if a mechanism is DP, one cannot tell whether any specific individual's data is included in the original dataset or not based on observing its output. For our problem, at a high-level, the canonical notion of DP requires that all outputs by the mechanism, including the payments it allocates to agents, are insensitive to each agent’s input. However, this is quite stringent since the payment to each agent is shared neither publicly nor with other agents. Thus, instead of the original DP, we consider one of its relaxations namely {\em joint differential privacy}~\cite{kearns2014mechanism}. 
    \begin{definition}[$\varepsilon$-Joint Differential Privacy~\cite{kearns2014mechanism}]
        \label{def:jdp}
        Consider a randomized mechanism $\mathcal{M}:\mathcal{D}^n\to \Theta\times \Pi^{n}$ with arbitrary response sets $\Theta, \Pi^n$. For each $i \in [n]$,  let $\mathcal{M}(\cdot)_{-i}=(\theta, \pi_{-i})\in \Theta\times \Pi^{n-1}$ denotes the portion of the mechanism’s output that
        is observable to outside observers and agents $j\neq i$. Then the mechanism $\mathcal{M}$ is $\epsilon$-jointly differentially private (JDP) if for every
        agent $i$, every dataset $D\in \mathcal{D}^n$ and every $D_i^{\prime}, D_i \in \mathcal{D}$  we have
        \begin{align*}
            \forall \mathcal{S}\subseteq \Theta\times \Pi^{n-1},
            \bb{P}\left(\mathcal{M}(D_i, D_{-i})_{-i}\in \mathcal{S}|(D_i,D_{-i})\right)\leq e^{\varepsilon}
            \bb{P}\left(\mathcal{M}(D_i^{\prime},D_{-i})_{-i}\in \mathcal{S}|(D_i^{\prime}, D_{-i})\right),
        \end{align*}
        where $D_{-i}\in \mathcal{D}^{n-1}$ is the dataset $D$ that excludes the $i$-th sample in $D$ and  $\pi_{-i}$ is the vector that comprises all payments excluding the payment of agent $i$. 
    \end{definition}
    In Definition~\ref{def:jdp} we assume that the private estimator $\theta$ computed by the mechanism $\mathcal{M}$ is a publicly observable output; in contrast, each payment $\pi_i$ can only be observed by agent $i$. Thus, from the view of each agent $i$, the mechanism output that is publicly released and that in turn might violate his/her privacy is $(\theta, \pi_{-i})$.
    
    In this paper, we further relax the definition of JDP by relaxing the requirement that the ratio between two output distributions is upper bounded for all pairs of datasets, to the requirement that the bounded ratio holds for likely dataset pairs. Specifically, motivated by the definition of random DP~\cite{hall2011random,rubinstein2017pain}, we consider
    {\em random joint differential privacy}:
    
    \begin{definition}[$(\varepsilon, \gamma)$-Random Joint Differential Privacy]
        \label{def:rjdp}
        Consider the same setting as in Definition~\ref{def:jdp}, we call a mechanism $\mathcal{M}$ preserves $(\varepsilon,\gamma)$-random joint differential privacy (RJDP), at privacy level $\varepsilon>0$ and confidence level $\gamma\in (0,1)$, if for every agent $i$, every dataset $D\in \mathcal{D}^{n}$ and every $D_i,D_{i}^{\prime}\in \mathcal{D}$ we have
        \begin{align*}
            \bb{P}[\forall \mathcal{S}\subseteq \Theta\times \Pi^{n-1},
            \bb{P}(\mathcal{M}(D_i, D_{-i})_{-i}\in\mathcal{S}|(D_i,D_{-i}))\leq e^{\varepsilon}
            \bb{P}(\mathcal{M}(D_i^{\prime},D_{-i})_{-i}\in\mathcal{S}|(D_i^{\prime}, D_{-i}))]\geq 1-\gamma.
        \end{align*}
        with the inner conditional probabilities take over the mechanism's randomization, and the outer probability takes over datasets $(D_i, D_{-i}), (D_i^{\prime}, D_{-i})$.
    \end{definition}
    
    %While strong $\epsilon$-JDP guarantee is ideal, utility may demand compromise.
    Note that there exists another relaxation of $\epsilon$-JDP called approximate JDP, or $(\epsilon, \delta)$-JDP, which is derived from $(\epsilon, \delta)$-DP. An $(\varepsilon, \delta)$-JDP mechanism on any dataset (including likely ones) may leak sensitive information on low probability responses, forgiven by the additive $\delta$ relaxation, while $(\epsilon, \gamma)$-RJDP offers an alternative relaxation, where on all but a small $\gamma$-proportion of unlikely dataset pairs, pure $\epsilon$-JDP holds. Similar to the approximate JDP, here we hope that $\gamma=o(\frac{1}{n})$.
    
    %The existence of priors and the independence of responses are only used to prove the accuracy of the learned model and truthfulness, but not to ensure any privacy guarantee. Our mechanism satisfies RJDP regardless of whether the assumptions hold; if they do, accuracy and truthfulness follow. Furthermore, although an agent can only manipulate $y_i$, both her/his manipulated response $\hat{y}_i$ and her/his features $\mathbf{x}_i$ are treated as “private” variables in our model, and both disclosures incur a privacy cost.

    \subsection{Utilities of Agents}
    \label{subsec:agents'_utilities}
    Based on the privacy definition, we now present the model on agents' utility. Here we adopt a similar assumption on the privacy cost as in the previous work \cite{cummings2015truthful,ghosh2011selling}. Specifically, for each agent~$i$, he/she has a privacy cost parameter $c_i$ and a privacy cost function $f_i(c_i, \varepsilon, \gamma)$ which measures the cost he/she incurs when his/her data is used in an $(\epsilon, \gamma)$-RJDP mechanism. Moreover, with payment $\pi_i$, we define agent $i$'s utility from reporting his/her data as $u_i=\pi_i-f_i(c_i,\varepsilon, \gamma)$. In this paper, following the previous work, we assume that all functions $f_i$ are bounded above by a function of $\epsilon, \gamma$ and $c_i$.
    \begin{assumption}
        \label{assump2:privacy_cost_function_bound}
        The privacy cost function of each agent satisfies
        \begin{align*}
            f_i(c_i,\varepsilon, \gamma)\leq c_iF(\varepsilon,\gamma).
        \end{align*}
        where $F(\varepsilon, \gamma)$ is an increasing function of both $\varepsilon$ and $\gamma$,
        and $F(\varepsilon, \gamma)\geq 0$ for all  $\varepsilon\in \bb{R}^{+}$.
    \end{assumption}
    Recall that $\varepsilon, \gamma$ are privacy parameters of the mechanism (see Definition~\ref{def:rjdp}). As larger values of $\varepsilon$ and $\gamma$ imply weaker privacy guarantee, which means the privacy cost of an agent becomes larger. Thus, it is natural to let $F$ be a component-wise increasing function. Note that in ~\cite{cummings2015truthful}, the authors consider the case where $\gamma=0$ and $F(\epsilon, \gamma)=\epsilon^2$. Thus, Assumption~\ref{assump2:privacy_cost_function_bound} can be considered as a generalization of their assumption.
    
    We also assume that each cost parameter $c_i$ is drawn independently from some distribution $\mathcal{C}$. Here we allow $c_i$ to be correlated with the data sample $D_i$. This is reasonable since for example, in a medical survey setting, if agent $i$ has a private value $y_i=1$ which means she/he is diagnosed as some disease, then she/he is probably more unwilling to truthfully report  the value, which implies $c_i$ is larger. However, we assume that an agent's cost coefficient $c_i$ does not provide any information about other agents:
    \begin{assumption}
        \label{assump3:privacy_cost_independence}
        Given $D_i, (D_{-i}, c_{-i})$ is conditionally independent of $c_i$:
        \begin{align*}
            p(D_{-i},c_{-i}| D_i,c_i)=
            p(D_{-i},c_{-i}|D_i,c_i^{\prime}) \quad \text{for all }
            D_{-i},c_{-i}, D_i, c_i,c_{i}^{\prime}. 
        \end{align*}
        where $c_{-i}$ is the collection of privacy costs excluding the privacy cost of agent $i$. 
    \end{assumption}
    
    In addition, we assume that the probability distribution of $c_i$ has exponential decay. Actually we can relax the assumption to polynomial decay and here the exponential decay assumption is only for simplicity.
    \begin{assumption}
        \label{assump4:privacy_cost_coefficient_tail}
        There exists some constant $\lambda>0$ such that the conditional distribution of privacy cost coefficient satisfies
        \begin{align*}
            \inf_{D_j}\bb{P}_{c_i\sim p(c_i|D_j)}(c_i\leq \tau)\geq 1-e^{-\lambda\tau}.
        \end{align*}
    \end{assumption}
    
    \subsection{Truthful Mechanisms}
    In this paper, we aim to design mechanisms that have the following properties: (1) truthful reporting is an equilibrium; (2) the private estimator of the outputs should be close to $\theta^*$; (3) the utilities for almost all agents are non-negative; (4) the payment budget required from the analyst to run the mechanism is small. We will quantify these properties using the notion of {\em Bayesian game}. A multiagent, one-shot, and simultaneous-move symmetric Bayesian game can model the outcome of agents' strategic reporting behavior. Formally, there are $n$ agents involved in the game. They privately observe their types $(\mathbf{x}_i, y_i, c_i)\stackrel{\mathrm{iid}}{\sim} \mathcal{F}\times \mathcal{G}_{\mathbf{x}_i}\times \mathcal{C}$. Each agent $i$ plays action $(\mathbf{x}_i, \hat{y}_i)$  and receives a real-valued payment $\pi_i$. And finally he/she receives utility $u_i=\pi_i-f_i(c_i,\varepsilon, \gamma)$. Let $\sigma_i$ denote agent $i$'s reporting strategy  (i.e., $\hat{y}_i=\sigma_i(y_i)$), $\sigma=(\sigma_1,\cdots, \sigma_n)$ denote the collection of all agents' strategies, and $\sigma_{-i}=(\sigma_1,\cdots, \sigma_{i-1},\sigma_{i+1},\cdots, \sigma_n)$ denote the collection of strategies except $\sigma_i$.
    \footnote{Note that throughout in the Bayesian game, the strategy spaces, the payoff functions, possible types, and the prior probability distribution are assumed to be common knowledge.}
    Based on this, we first quantify the property (1) by Bayesian Nash equilibrium.
    
    \begin{definition}[$\eta$-Bayesian Nash equilibrium]
        A reporting strategy profile $\sigma=(\sigma_1,\cdots, \sigma_n)$ forms an $\eta$-Bayesian Nash equilibrium if for every agent $i$, $D_i$ and $c_i$, and for any other reporting strategy $\sigma^{\prime}_i\neq \sigma_i$,
        \begin{align*}
            &\quad \bb{E}_{D_{-i, c_{-i}}\sim p(D_{-i}, c_{-i}|D_i, c_i)}[u_i(\sigma_i(D_i, c_i), \sigma_{-i}(D_{-i}, c_{-i}))] \\
            &\geq \bb{E}_{D_{-i}, c_{-i}\sim p(D_{-i}, c_{-i}|D_i, c_i)}[u_i(\sigma^{\prime}_i(D_i, c_i),\sigma_{-i}(D_{-i}, c_{-i}))]-\eta.
        \end{align*}
    \end{definition}
    
    The positive value $\eta$ quantifies at most how much additional expected payment an agent can receive if she/he changes her/his reporting strategy.  As we want all agents to truthfully report their data, we require the payment rule to keep $\eta$ as small as possible. In this paper we consider the following threshold strategy. We will show that if all agents follow the threshold strategy with some common positive value $\tau$, then such a strategy profile achieves an $\eta$-Bayesian Nash equilibrium.
    
    \begin{definition}[Threshold strategy]
        Define the threshold strategy $\sigma_{\tau}$ as follows:
        \begin{align*}
            \hat{y}_i=\sigma_{\tau}(\mathbf{x}_i, y_i, c_i)=
            \begin{cases}
                y_i, \quad \text{if}\quad  c_i\leq \tau, \\
                \text{arbitrary value in $\mathcal{Y}$}, \quad \text{if} \quad  c_i>\tau.
            \end{cases}
        \end{align*}
    \end{definition}
    
    To link truthful reporting strategy threshold $\tau$ and privacy cost coefficients $c_i$, following~\cite{cummings2015truthful}, we use the following definition.
    
    \begin{definition}
        \label{def:alpha_beta}
        Fix a probability density function $p(c)$ of privacy cost parameter,  and let
        \begin{align*}
            &\tau^1_{\alpha,\beta}=\inf\{\tau>0: \bb{P}_{(c_1,\cdots,c_n)\sim p^{n}}(\#\{i:c_i\leq \tau\}\geq (1-\alpha)n)\geq 1-\beta\},\\
            & \tau_{\alpha}^2 = \inf\{\tau>0: \inf_{D_i}\bb{P}_{c_j\sim p(c|D_i)}(c_j\leq \tau)\geq 1-\alpha\}.
        \end{align*}
        Define $\tau_{\alpha,\beta}$ as the larger of these two thresholds:
        $\tau_{\alpha,\beta}=\max\{\tau_{\alpha,\beta}^1,\tau_{\alpha}^2\}.$
    \end{definition}
    
    Note that $\tau^1_{\alpha,\beta}$ is such a threshold that with probability at least $1-\beta$, at least $1-\alpha$ fraction of agents have cost coefficient $c_i\leq \tau_{\alpha,\beta}$. And $\tau_{\alpha}^2$ is such a threshold that conditioned on his/her own dataset $D_i$, each agent $i$ believes that with probability $1-\alpha$ any other agent $j$ has cost coefficient $c_j\leq \tau_{\alpha,\beta}$.
    
    For property (2), we use the square of $\ell_2$-norm distance between the private estimator $\bar{\theta}^P$ and the true parameter $\theta^*$.
    
    \begin{definition}[$\eta$-accurate]
        We call the mechanism is $\eta$-accurate if its output $\bar{\theta}^P$ satisfies $\bb{E}[\|\bar{\theta}^P-\theta^*\|_2^2]\leq \eta$.
    \end{definition}
    
   To satisfy property (3), we should make payments high enough to compensate privacy cost.
    
    \begin{definition}[Individual rationality]
        Let $u_i$ denote the utility agent $i$ receives. A mechanism is individually rational if  $\bb{E}[u_i]\geq 0$ for every agent $i$.
    \end{definition}
    
    We also concern on the total amount of payment budget required by the analyst to run the mechanism, and want it tend to be zero  when the number of agents increases.  
    
    \begin{definition}[Asymptotically small budget]
        An asymptotically small budget is such that $\mathcal{B}=\sum_{i=1}^n \bb{E}[\pi_i]=o(1)$ for all realizable $D=\{(\mathbf{x}_i, y_i)\}_{i=1}^n$.
    \end{definition}
    
    \section{Sub-Gaussian Case for Generalized Linear Models}
    \label{sec:GLM}
    In this section we consider generalized linear models~\eqref{def:glm_pdf} with sub-Gaussian covariates.
    
    \begin{definition}[Sub-Gaussian random variable]
        A zero-mean random variable $X\in \bb{R}$ is said to be sub-Gaussian with variance  $\sigma^2$ ($X\sim \mathrm{subG}(\sigma^2)$) if its moment generating function satisfies $\bb{E}[\exp(tX)]\leq \exp(\frac{\sigma^2t^2}{2})$ for all $t>0$.
    \end{definition}
    
    \begin{definition}[Sub-Gaussian random vector]
        A zero mean random vector $X\in \bb{R}^d$ is said to be sub-Gaussian with variance $\sigma^2$ ($X\sim \mathrm{subG}_d(\sigma^2)$) if $\langle X, u\rangle$ is sub-Gaussian with variance $\sigma^2$ for any unit vector $u\in \mathbb{R}^d$.
    \end{definition}
    
    The class of sub-Gaussian random variables is quite large. It includes bounded random variables and Gaussian random variables, and it enjoys strong concentration properties.
    
    \begin{lemma}[\cite{vershynin2018high}]\label{lem:subG_tail}
        If $X\sim \mathrm{subG}(\sigma^2)$, then for any $t>0$, it holds that
        $\bb{P}(|X|>t)\leq 2\exp(-\frac{t^2}{2\sigma^2})$.
    \end{lemma}
    
    \begin{lemma}[\cite{vershynin2018high}]\label{lem:subG_bound}
        For a sub-Gaussian vector $X\sim \text{subG}_d(\sigma^2)$, with probability at least $1-\delta$ we have $\|X\|_2\leq 4\sigma \sqrt{d\log \frac{1}{\delta}}$.
    \end{lemma}
    
    We make the following assumptions used throughout this section.
    
    %    The first one is that each $\mathbf{x}_i$ is sub-Gaussian and the $\ell_\infty$-norm and $\ell_2$-norm of the covariance matrix of $\mathbf{x}_i$ are bounded below by some constant.
    
    \begin{assumption}
        \label{assump5:subG}
        The covariates  $\mathbf{x}_1, \mathbf{x}_2, \cdots, \mathbf{x}_n \in \mathbb{R}^d$ are i.i.d. (zero-mean) sub-Gaussian random vectors with variance $\frac{\sigma^2}{d}$ with $\sigma=O(1)$, Moreover, the covariance matrix $\Sigma$ of $\mathbf{x}_i$ satisfies that $\|\Sigma\|_\infty\geq \kappa_\infty$ and $\|\Sigma\|_2 \geq \kappa_2$ for constants $\kappa_{\infty}, \kappa_2=\Theta(1)$, {\em i.e.},  $\forall w\in \bb{R}^d$, $\|\Sigma w\|_{\infty}\geq \kappa_{\infty}\|w\|_{\infty}$ and $\|\Sigma w\|_2\geq \kappa_2\|w\|_2$. We also assume that $y_i$ have finite fourth moment $R:=\bb{E}[y_i^4]=O(1)$.~\footnote{For simplicity, here we assume the variance proxy as $\frac{\sigma^2}{d}$ is to make $\|x_i\|_2$ bounded by a constant so that we can compare with the previous work on linear regression with bounded covariates. We can extend all of our results to the general $\sigma^2$ case with additional factor of $\text{Poly}(d)$ in the upper bounds of our results. } We focus on the low dimension case where $n=\tilde{\Omega}(d)$, where $\tilde{\Omega}$ omits the term of $\log n$.~\footnote{It is also notable that for all the constants, Big-$O$ and Big-$\Omega$ notations of  in this paper we omit the terms of $\sigma, R, \kappa_2, \kappa_\infty, \lambda_{\max}$ as we assume they are constants, where $\lambda_{\max}$ is the largest eigenvalue of $\Sigma$. See the proofs in Appendix for the full version of the results.}
    \end{assumption}
    Note that by Lemma~\ref{lem:subG_bound}, with high probability each $\|\mathbf{x}_i\|_2$ is upper bounded by a constant. This can be thought as a generalization of~\cite{cummings2015truthful} which assumes all $\|\mathbf{x}_i\|_2$ are  bounded by a constant. However,~\cite{cummings2015truthful} also assumes that each $y_i$ is also bounded where here we assume that it only has finite fourth moment. %Such assumption is natural as we can show that when the function $A'(\cdot)$ and $A''(\cdot)$ is bounded, then such the assumption holds.
    
    \subsection{Main Idea}
    \label{subsec:main_idea}
    Before showing our method, we first go back to estimating $\theta^*$ without privacy constraint and manipulating the response. Given $n$ samples $\{(\mathbf{x}_i, y_i)\}_{i=1}^{n}$, the  maximum likelihood estimator of $\theta^{*}$ based on the probability distribution~\eqref{def:glm_pdf} is given by
    \begin{align*}
        \tilde{\theta}\in \arg\max_{\theta} \prod_{i=1}^{n}p(y_i | \mathbf{x}_i;\theta) =\arg\min_{\theta} -\frac{1}{n}\theta^TX^Ty+\frac{1}{n}\textbf{1}^TA(X\theta),
        \numberthis\label{mle}
    \end{align*}
    where $X=(\mathbf{x}_1^T, \cdots, \mathbf{x}_n^T)^T \in \mathbb{R}^{n\times d}, y=(y_1,\cdots, y_n)^T\in \mathbb{R}^{n}$ and $A(X\theta) \in \mathbb{R}^n$ with $[A(X\theta)]_j=A(\mathbf{x}_j^T \theta)$. Since $\tilde{\theta}$ is the maximizer of likelihood function~\eqref{mle}, motivated by \cite{yang2015closed}, it must satisfies the stationary condition:
    \begin{align*}
        X^Ty=X^T\nabla A(X\tilde{\theta}), 
    \end{align*}
   where $\nabla A(\eta)\equiv (A^{\prime}(\eta_1), \cdots, A^{\prime}(\eta_n))^T\in \mathbb{R}^n$ for any $\eta\in \bb{R}^n$.  Intuitively, it means that $\nabla A(X\hat{\theta})\approx y$ which implies $X\tilde{\theta}\approx [\nabla A]^{-1}(y)$, where $[\nabla A]^{-1}(y)\equiv ((A^{\prime})^{-1}(y_1), \cdots, (A^{\prime})^{-1}(y_n))$. However, the challenge here is that the function $(A^{\prime})^{-1}(\cdot)$ may not be well defined on every point of $\mathcal{Y}$.  In fact the function $A^{\prime}(\cdot)$ is only onto the interior $\mathcal{M}^{o}$ of the response moment polytope $\mathcal{M}$, which is defined as  $\mathcal{M}:=\{\mu:\mu=\bb{E}_p[y], \text{for some distribution $p$ over $y\in \mathcal{Y}$}\}$~\cite{wainwright2008graphical}. Thus to make it well-defined we should project each $y_i$ onto $\mathcal{M}^{o}$ first. However, as $\mathcal{M}^{o}$ is an open set, the projection operator may not be well-defined. Thus, we use a closed subset of $\mathcal{M}^{o}$ instead.  In this paper, for different GLM models, we construct a different closed subset $\mathcal{\bar{M}}$ of the interior $\mathcal{M}^{o}$. The projection operator $\Pi_{\mathcal{\bar{M}}}(\cdot)$ is defined as $\Pi_{\mathcal{\bar{M}}}(y_i)=\arg\min_{\mu\in\mathcal{\bar{M}}}|y_i-\mu|$ for any variable $y_i\in \mathcal{Y}$, and $[\Pi_{\mathcal{\bar{M}}}(y)]_i=\Pi_{\mathcal{\bar{M}}}(y_i)$ for any vector $y\in \bb{R}^n$. After projecting each $y_i$, we can approximate $\tilde{\theta}$ via the least square method on $(X, [\nabla A]^{-1}(\Pi_{\bar{\mathcal{M}}}(y)))$. In total we have 
    \begin{align*}
        \tilde{\theta}\approx (\frac{X^TX}{n})^{-1}\frac{X^T[\nabla A]^{-1}(\Pi_{\bar{\mathcal{M}}}(y))}{n}.
        \numberthis\label{estimator}
    \end{align*}
    From~\eqref{estimator} we can see it is sufficient for the $i$-th agent to share $x_i, y_i$ and the analyst computes $\mathbf{x}_i^T\mathbf{x}_i$ and $\mathbf{x}_i^T(A')^{-1}(\Pi_{\bar{\mathcal{M}}}(y_i))$. To achieve RJDP, by the basic mechanism in DP one direct approach is to add noise to the term  $(\frac{X^TX}{n})^{-1}\frac{X^T[\nabla A]^{-1}(\Pi_{\bar{\mathcal{M}}}(y) }{n}$, where the magnitude should be proportional to the sensitivity of the term. However, the challenge here is that the sensitivity of the RHS term in~\eqref{estimator} maybe unbounded with constant probability. To be more concrete, there are two terms $(\frac{X^TX}{n})^{-1}$ and $\frac{X^T[\nabla A]^{-1}(\Pi_{\bar{\mathcal{M}}}(y) }{n}$. The sensitivity of the term $(\frac{X^TX}{n})^{-1}$ is bounded with high probability due to Lemma~\ref{lem:subG_bound}. However, the main issue is, as we only assume $y_i$ has bounded fourth moment, the term $[\nabla A]^{-1}(\Pi_{\bar{\mathcal{M}}}(y)$ could be unbounded with high probability (such as Poisson regression). And this could cause the sensitivity of the RHS term in~\eqref{estimator} be unbounded. To overcome the challenge, here we will further conduct a clipping step, that is, we shrink each $y_i$ into an bounded interval $[-\tau_2, \tau_2]$ for some finite positive value $\tau_2$:
    \begin{align*}
        \widetilde{y}_i:=\mathrm{sgn}(y_i)\min\{|y_i|, \tau_2\}.
        \numberthis\label{shrink_y}
    \end{align*}
    In total, our non-private estimator will be
    \begin{align*}
        \hat{\theta}(D)=(X^TX)^{-1}X^T(\nabla A)^{-1}(\proj(\widetilde{y})).
        \numberthis\label{est_glm}
    \end{align*}
    Later we will show that with high probability the $\ell_2$-norm sensitivity of~\eqref{est_glm} is bounded. Thus, we can add noise to $\hat{\theta}(D)$ to make it private: $\hat{\theta}^P(D)=\hat{\theta}(D)+\text{noise}$. Since we assume $\|\theta^*\|_2\leq \tau_\theta$, we need to project $ \hat{\theta}^P(D)$ onto a $\ell_2$-norm ball:
    \begin{align*}
        \bar{\theta}^P(D)=\Pi_{\tau_\theta}( \hat{\theta}^P(D)),
        \numberthis\label{est_glm2}
    \end{align*}
    where $\Pi_{\tau_\theta}(v)=\arg\min_{v'\in{\mathbb{B}(\tau_\theta)}}\|v'-v\|^2_2$ and ${\mathbb{B}(\tau_\theta)}$ is the closed $\ell_2$-norm ball with radius $\tau_{\theta}$ and  centers at the origin.
    
    Previously we focused on the privacy and accuracy. In the following we will consider the payment rule. The analyst should pay each agent strategically. If the analyst knows the ground-truth after collecting the reports, she/he can pay each agent according to how well the reports are aligned with the ground-truth, e.g., by using $\ell_2$ norm as the distance metric for relatedness. However, in our setting, the data is unverifiable, which means we do not have ground truth  as reference. To deal with this problem, we adopt the \textit{peer prediction method}~\cite{miller2005eliciting,dasgupta2013crowdsourced,shnayder2016informed,agarwal2020peer,chen2020truthful}, which extracts information from peers' reports used for reference. In other words, each agent will receive higher payment if her/his report is more consistent with the statistical model estimated by using other agents' reports. Since we assume that all data are generated by the same statistical model, the peer prediction method intuitively encourages truthful reporting if most agents report their data truthfully. There are many ways to quantify the relatedness between each agent's reports and her/his peers' reports, e.g., point-wise mutual information~\cite{kong2018water}, delta matrices~\cite{agarwal2020peer}, the Brier scoring rule~\cite{ghosh2014buying}. Here we adopt the rescaled Brier score rule. Formally, the analyst uses payment rule
    \begin{align*}
        B_{a_1,a_2}(p,q)=a_1-a_2(p-2pq+q^2),
        \numberthis\label{brier_score_rule}
    \end{align*}
    where $a_1, a_2>0$ are parameters to be determined, $q$ is the prediction of agent $i$'s response given her/his reports, and $p$ is the prediction of agent $i$'s response given her/his feature vector and her/his peers' reports. Note that $B_{a_1,a_2}(p,q)$ is a strictly concave function of $q$ which is maximized at $q=p$, which means the prediction of agent $i$'s response given her/his information is aligned with the one given peers' information. In GLMs, for agent $i$, since $\bb{E}[y_i|\mathbf{x}_i, \theta^{*}]=A^{\prime}(\langle\mathbf{x}_i, \theta^{*}\rangle)$, it is natural to let $p=A^{\prime}(\langle\mathbf{x}_i, \bar{\theta}^P(\hat{D}^b)\rangle)$ and $q=A^{\prime}(\langle\mathbf{x}_i, \bb{E}_{\theta\sim p(\theta|\hat{D}_i)}[\theta]\rangle)$, where $\bar{\theta}^P(\hat{D}^b)$ is the private estimator on a  dataset $\hat{D}^b$ that does not include $\hat{D}_i$, and $p(\theta|\hat{D}_i)$ is the posterior distribution of $\theta$ after the analyst receives $\hat{D}_i$.
    
    Based on the previous analysis, we formalize our Mechanism~\ref{mech:GLM}. Note that instead of using agents' original data, we use the reported data (which may contain manipulated responses) to obtain the estimator.  In order to eliminate dependency, we need to partition the dataset into two subgroups $\hat{D}^0$ and $\hat{D}^1$. To calculate the payment for each agent $i$ in group $b\in \{0, 1\}$, we use $\hat{D}^{1-b}$ to estimate $\theta^{*}$, and then use the estimator and her/his feature vector $\mathbf{x}_i$ to predict the response.

    \subsection{Theoretical Analysis}
    \label{subsec:theoretical_analysis}
    Before showing our results, we first list some notations for later use. By Assumption~\ref{assump5:subG} and Lemma~\ref{lem:subG_bound}, with probability at least $1-n^{-\Omega(1)}$, it holds that $\|\mathbf{x}_i\|_2\leq C\sigma \sqrt{\log n}$ for all $i\in [n]$ with sufficiently large $C>0$. Denote $\tau_1=C \sigma \sqrt{\log n}$, then $|\langle \mathbf{x}_i, \theta^*\rangle|\leq\tau_1\tau_{\theta}$ for all $i\in [n]$.  The following notations correspond to the upper bounds of some functions on some closed sets.
    \begin{align*}
        &\mathcal{M}^{\prime}:=\{\mu:\mu=A^{\prime}(a), a\in [-\tau_{\theta}\tau_1, \tau_{\theta}\tau_1]\},\\
        &\kappa_{A,0}:=\max_{a\in \mathcal{M}^{\prime}\cup \bar{\mathcal{M}}}|[(A^{\prime})^{-1}]^{\prime}(a)|, \quad
        \kappa_{A,1}:=\max_{a\in \bar{\mathcal{M}}\cap [-\tau_2, \tau_2]}|(A^{\prime})^{-1}(a)|,\\
        &\kappa_{A,2}:=\max_{a\in[-\tau_{\theta}\tau_1,\tau_{\theta}\tau_1]}|A^{\prime\prime}(a)|,\quad M_A:=\max_{a\in [-\tau_{\theta}\tau_1,\tau_{\theta}\tau_1]}|A^{\prime}(a)|,\\
        &\varepsilon_{\bar{\mathcal{M}}}:=\max_{y_i\in \mathcal{Y}\cap [-\tau_2,\tau_2]}|y_i-\proj(y_i)|,
    \end{align*}
    where $\tau_2$ is the threshold value in~\eqref{shrink_y} and $\bar{\mathcal{M}}$ is the closed set in~\eqref{est_glm}. Note that all these parameters depend on the link function $A$, which varies for different specific models. Thus, here we cannot assume they are constants. In the following we will always assume Assumptions~\ref{assump1:theta&lowD}-\ref{assump5:subG} hold.
    \begin{Mechanism}[htbp]
        \label{mech:GLM}
        \caption{Private Generalized Linear Models Mechanism}
        Ask all agents to report their data $\hat{D}_1, \cdots, \hat{D}_n$\;
        Randomly partition agents into two groups, with respective data pairs $\hat{D}^0, \hat{D}^1$\;
        Compute estimators $\hat{\theta}(\hat{D}), \hat{\theta}(\hat{D}^0),  \hat{\theta}(\hat{D}^1)$ according to~\eqref{est_glm} on $\hat{D}, \hat{D}^0$ and $\hat{D}^1$ respectively\;
        Compute  estimator sensitivity $\Delta_n, \Delta_{n/2}$, and set differential privacy parameter $\varepsilon$\;
        Draw $v\in \bb{R}^d$ according to distribution $p(v)\propto \exp(-\frac{\varepsilon}{\Delta_n}\|v\|_2)$, and independently draw $v_{0},v_{1}\in \bb{R}^d$ according to distribution
        $p(v)\propto \exp(-\frac{\varepsilon}{\Delta_{n/2}}\|v\|_2)$\;
        Add noise: $\hat{\theta}^P(\hat{D})=\hat{\theta}(\hat{D})+v,  \hat{\theta}^{P}(\hat{D}^b)=\hat{\theta}({\hat{D}^b})+v_{b}$ for $b=0,1$\;
        Compute private estimators $\bar{\theta}^P(\hat{D})=\Pi_{\tau_\theta}(\hat{\theta}^P(\hat{D}))$ and $\bar{\theta}^{P}(\hat{D}^b)=\Pi_{\tau_\theta}(\hat{\theta}^{P}(\hat{D}^b))$ for $b=0,1$\;
        Set parameters $a_1, a_2$, and compute payments to each agent $i$:  if agent $i$'s is in group $1-b$, then he will receive payment
        \begin{align*}
            \pi_i=B_{a_1,a_2}\left(A^{\prime}(\langle  \mathbf{x}_i,\bar{\theta}^{P}(\hat{D}^{b})\rangle),A^{\prime}(\langle \mathbf{x}_i, \bb{E}_{\theta\sim p(\theta|\hat{D}_i)}[\theta] \rangle)\right).
        \end{align*}
    \end{Mechanism}
    \begin{lemma}[Sensitivity]
        \label{lem:sensitivity_glm}
        With probability at least $1-C_1n^{-\Omega(1)}$ the $\ell_2$-norm sensitivity of $\hat{\theta}(D)$ computed by~\eqref{est_glm} satisfies
        \begin{align*}
            \max_{D\sim D'}\|\hat{\theta}(D)-\hat{\theta}(D')\|_2\leq \Delta_n= C_0\kappa_{A, 1} \frac{\sqrt{d\log n}}{\sqrt{n}},
        \end{align*}
        where $ C_0, C_1>0$ are  constants.  For later use, we denote  $\gamma_n=C_1n^{-\Omega(1)}$ as the failure probability.
    \end{lemma}
    
    \begin{lemma}[Accuracy of the non-private estimator]
        \label{lem:accuracy_glm}
        With probability at least $1-O(n^{-\Omega(1)})$ one has
        \begin{align*}
            \|\hat{\theta}(D)-\theta^{*}\|_2\leq \lambda_n \sqrt{\frac{\log n}{n}},
        \end{align*}
        where $ \lambda_n:=\tilde{O}({\kappa_{A,0}}(\sqrt{\kappa_{A,2}+\frac{1}{\tau_2^2}}+(M_A+\tau_2)\sqrt[4]{\frac{1}{n}}+\varepsilon_{\bar{\mathcal{M}}})$ and the Big-$\tilde{O}$ notation omits the term of $\mathrm{Poly}(\log n)$.
    \end{lemma}
    
    The previous lemma indicates that the closed-form estimator~\eqref{est_glm} on the original dataset is consistent.  It is notable that its convergence rate may not be as fast as $\tilde{O}(\sqrt{\frac{1}{n}})$ since $\lambda_n$ has different growth rates in different specific models.
    \begin{theorem}[Privacy]
        \label{thm:privacy_glm}
        Mechanism~\ref{mech:GLM} satisfies $(2\varepsilon,\gamma_n+2\gamma_{n/2})$-random joint differential privacy, where $\gamma_n=C_1n^{-\Omega(1)}$ is in Lemma~\ref{lem:sensitivity_glm}.
    \end{theorem}
    
    \begin{theorem}[Truthfulness]
        \label{thm:truthfulness_glm}
        Fix a privacy parameter $\varepsilon$, a participation goal $1-\alpha$ and a desired confidence parameter $\beta$ in Definition~\ref{def:alpha_beta}. Then with probability at least $1-\beta-O(n^{-\Omega(1)})$, the symmetric threshold strategy $\sigma_{\tau_{\alpha,\beta}}$ is an $\eta$-Bayesian Nash equilibrium in Mechanism~\ref{mech:GLM} with
        \begin{align*}
            \eta =\tilde{O}\left(a_2\kappa_{A,2}^2(\alpha^2 \kappa^2_{A, 1} {{{nd}}} +{\frac{\lambda^2_n}{n}}+\frac{\kappa^2_{A,1}d^3}{n\epsilon^2})+\tau_{\alpha,\beta}F(2\varepsilon, \gamma_n+2\gamma_{n/2})\right),
        \end{align*}
        where $\gamma_n=C_1n^{-\Omega(1)}$ is in Lemma~\ref{lem:sensitivity_glm}, $\lambda_n$ is in Lemma~\ref{lem:accuracy_glm}, $a_2$ is in \eqref{brier_score_rule}, $\tau_{\alpha,\beta}$ is in Definition~\ref{def:alpha_beta} and function $F$ is in Assumption~\ref{assump2:privacy_cost_function_bound}.
    \end{theorem}
    
    \begin{theorem}[Accuracy]
        \label{thm:accuracy_glm}
         Fix a privacy parameter $\varepsilon$, a participation goal $1-\alpha$ and a desired confidence parameter $\beta$ in Definition~\ref{def:alpha_beta}. Then  under the symmetric threshold strategy $\sigma_{\tau_{\alpha,\beta}}$, the output $\bar{\theta}^P(\hat{D})$ of  Mechanism~\ref{mech:GLM} satisfies that with probability at least $1-\beta-O(n^{-\Omega(1)})$,
        \begin{align*}
            \bb{E}[\|\bar{\theta}^P(\hat{D})-\theta^{*}\|_2^2]\leq
            \tilde{O}\left ( \alpha^2 \kappa^2_{A, 1}nd+ \frac{\kappa^2_{A, 1} {{d^3}}}{\varepsilon^2 n} + {\frac{\lambda^2_n}{n}} \right),
        \end{align*}
        where $\lambda_n$ is in Lemma~\ref{lem:accuracy_glm}.
    \end{theorem}

    \begin{theorem}[Individual rationality]
        \label{thm:rationality_glm}
        With probability at least $1-\beta-O(n^{-\Omega(1)})$, Mechanism~\ref{mech:GLM} is individually rational for all agents with cost coefficients $c_i\leq \tau_{\alpha,\beta}$ as long as
        \begin{align*}
            a_1\geq a_2(M_A+3M_A^2)+\tau_{\alpha,\beta}F(2\varepsilon, \gamma_n+\gamma_{n/2})
        \end{align*}
        regardless of the reports from agents with cost coefficients above $\tau_{\alpha,\beta}$,  where $\gamma_n=C_1n^{-\Omega(1)}$ is  in Lemma~\ref{lem:sensitivity_glm}, $a_1, a_2$ are in~\eqref{brier_score_rule} and $\lambda_n$ is in Lemma~\ref{lem:accuracy_glm}.
    \end{theorem}
    
    \begin{theorem}[Budget]
        \label{thm:budget_glm}
        With probability at least $1-\beta-O(n^{-\Omega(1)})$, the total expected budget $\mathcal{B}:=\bb{E}[\sum_{i=1}^{n}\pi_i]$ required by the analyst to run Mechanism~\ref{mech:GLM} under threshold equilibrium strategy $\sigma_{\tau_{\alpha,\beta}}$ satisfies
            $$\mathcal{B}\leq n(a_1+a_2(M_A+M_A^2)),$$ where $a_1, a_2$ are in~\eqref{brier_score_rule}. 
    \end{theorem}
       \vspace{-0.2in}
    \subsection{Implementation for Some Specific Models}
       \vspace{-0.1in}
    \label{subsec:examples}
    In this section we will  apply our framework to three canonical models in GLM: linear regression, logistic regression and Poisson regression. Based on our previous results, to provide appropriate design of computation and payment scheme  it is sufficient to construct $\bar{\mathcal{M}}$, specify the growth rates of  $\{\kappa_{A,i}\}_{i=0}^2, M_A, \varepsilon_{\bar{\mathcal{M}}}$, and set suitable parameters including $\alpha, \beta, \varepsilon, a_1, a_2, \tau_2$. In this section, we suppose that the privacy cost dominated function $F(\varepsilon, \gamma)$ in Assumption~\ref{assump2:privacy_cost_function_bound} satisfies $F(\varepsilon, \gamma)= (1+\gamma)\varepsilon^4$ for simplicity. It is notable that functions of $F$ can also be other functions. Moreover, for some specific models  we may allow more relaxed assumptions on the dependency of $\epsilon$ and $\gamma$ in $F(\epsilon, \gamma)$. 
       \vspace{-0.1in}
    \subsubsection{Linear Regression}\label{subsec:linear}
       \vspace{-0.1in}
    \begin{example}
        Consider the (Gaussian) linear regression model $y=\langle \theta^{*}, \mathbf{x}\rangle+\zeta$, where random variables $\mathbf{x}$ and $\zeta$ are independent, and $\zeta\sim \mathcal{N}(0,\sigma^2)$. Then conditioned on $\mathbf{x}$, the response $y$ follows the distribution $p(y|\mathbf{x};\theta^{*})
        =\frac{1}{\sqrt{2\pi}\sigma}\exp\{-\frac{(y-\langle\mathbf{x}, \theta^{*}\rangle)^2}{2\sigma^2}\}
        =\exp\{\frac{y\langle\mathbf{x}, \theta^{*}\rangle-\frac{1}{2}\langle\mathbf{x}, \theta^{*}\rangle^2}{\sigma^2}-\frac{1}{2}(\frac{y^2}{\sigma^2}+\log (2\pi \sigma^2))\}$. Thus, $A(a)=\frac{1}{2}a^2$, $\phi=\sigma^2$, and $c(y, \phi)=-\frac{1}{2}(\frac{y^2}{\sigma^2}+\log (2\pi\sigma^2))$.
    \end{example}

    \begin{corollary}\label{cor:glm_linear}
        %        Suppose that $F(\varepsilon, \gamma)\sim \varepsilon^4\gamma$ when $\varepsilon\to 0, \gamma\to 0$. 
        For any $\delta\in (\frac{1}{4}, \frac{1}{3})$ and $c>0$, we set $\bar{\mathcal{M}}=\bb{R}$ ( we have
        $\Pi_{\mathcal{\bar{M}}}(\widetilde{y}_i)=\widetilde{y}_i$), 
        $\varepsilon=n^{-\delta}$, $\tau_2=n^{\frac{1-3\delta}{2}}$ $\alpha=\Theta(n^{-3\delta})$, $\beta=\Theta(n^{-c})$, $a_2=O(n^{-4\delta})$, $a_1=a_2(M_A+3M_A^2)+\tau_{\alpha,\beta}F(2\varepsilon, \gamma_n+2\gamma_{n/2})$. Then 
        the output of Mechanism~\ref{mech:GLM} satisfies $(O(n^{-\delta}), O(n^{-\Omega(1)}))$-RJDP. Moreover, with probability at least $1-O(n^{-\Omega(1)})$, it holds that: \footnote{Note that for clearness and to be consistent with the previous results \cite{cummings2015truthful} here we omit the term of $\text{Poly}(d)$, the same for all the other corollaries in this paper.} 
        \begin{itemize}
            \item the symmetric threshold strategy $\sigma_{\tau_{\alpha,\beta}}$ is a $\widetilde{O}(n^{-4\delta})$-Bayesian Nash equilibrium for a $1-O(n^{-3\delta})$ fraction of agents to truthfully report their data;
            \item the private estimator $\bar{\theta}^P(\hat{D})$ is $\widetilde{O}(n^{-\delta})$-accurate;
            \item it is individually rational for a $1-O(n^{-3\delta})$ fraction of agents to participate in the mechanism;
            \item the total expected budget required by the analyst is $\widetilde{O}(n^{-4\delta+1})$.
        \end{itemize}
    \end{corollary}
    
    \begin{remark}
        In~\cite{cummings2015truthful}, the authors also study the problem of truthful linear regression. Specifically, they show that under the assumption of $F(\epsilon, \gamma)=\epsilon^2$ it is possible to design an $o(\frac{1}{\sqrt{n}})$-JDP mechanism that is an $o(\frac{1}{n})$-approximate Bayes Nash equilibrium, $o(1)$-accurate, individually rational for $(1-o(1))$ fraction of truthful agents and needs $o(1)$ budgets. In comparison, here we need stronger dependency on $\epsilon$ in the function $F$ and our algorithm can only guarantee $o(\frac{1}{\sqrt[4]{n}})$-RJDP. However, it is notable that $o(\frac{1}{\sqrt[4]{n}})$-RJDP is still in the extremely high privacy regime as in practice $\epsilon=0.1-0.5$ is enough to preserve privacy. For the dependency of $\epsilon$ in $F$, it is notable that~\cite{cummings2015truthful} need to assume $y_i$ is bounded where here we relax it to the case where it only has finite fourth moment. Thus, their results are incomparable with ours. 
    \end{remark}
   \vspace{-0.2in}
    \subsubsection{Logistic Regression}
       \vspace{-0.1in}
    \begin{example}
       Here the response $y\in \mathcal{Y}\equiv\{-1,1\}$. Let $p:=\bb{P}(y=1|\mathbf{x}_i,\theta^{*})$, then the conditional distribution of $y$ can be written as $p^{\frac{y+1}{2}}(1-p)^{\frac{1-y}{2}}=\exp\{\frac{y}{2}\log\frac{p}{1-p}+\frac{1}{2}\log p(1-p)\}$. If we set $\langle\mathbf{x}_i, \theta^{*}\rangle=\frac{1}{2}\log \frac{p}{1-p}$, then $p=\frac{e^{\langle\mathbf{x}_i, \theta^{*}\rangle}}{e^{\langle\mathbf{x}_i, \theta^{*}\rangle}+e^{-\langle\mathbf{x}_i, \theta^{*}\rangle}}$ and the above distribution is equal to $\exp\{y\langle\mathbf{x}_i, \theta^{*}\rangle-\log (\exp(-\langle\mathbf{x}_i, \theta^{*}\rangle)+\exp(\langle\mathbf{x}_i, \theta^{*}\rangle))\}$. Hence, here $A(a)=\log(e^{-a}+e^{a})$, $\phi=1$, and $c(y,\phi)=0$.
    \end{example}

    \begin{corollary}\label{cor:glm_logistic}
        %        Suppose that $F(\varepsilon, \gamma)\sim \varepsilon^4\gamma$ when $\varepsilon\to 0, \gamma\to 0$. 
        For any $\delta\in (\frac{1}{4}, \frac{1}{2})$ and $c>0$, we set $\bar{\mathcal{M}}=[-1+\varepsilon^{\prime}, 1-\varepsilon^{\prime}]$ for  $\varepsilon^{\prime}=2n^{-\delta}$ (we have $\Pi_{\mathcal{\bar{M}}}(\widetilde{y}_i)=\widetilde{y}_i(1-2n^{-\delta})$), $\varepsilon=n^{-\delta}$, $\tau_2=1$, $\alpha=\Theta(n^{-3\delta})$, $\beta=\Theta(n^{-c})$, $a_2=O(n^{-4\delta})$, $a_1=a_2(M_A+3M_A^2)+\tau_{\alpha,\beta}F(2\varepsilon, \gamma_n+2\gamma_{n/2})$. Then 
        the output of Mechanism~\ref{mech:GLM} satisfies $(O(n^{-\delta}), O(n^{-\Omega(1)}))$-RJDP. And with probability at least $1-O(n^{-\Omega(1)})$, it holds that:
        \begin{itemize}
            \item the symmetric threshold strategy $\sigma_{\tau_{\alpha,\beta}}$ is a $\widetilde{O}(n^{-4\delta})$-Bayesian Nash equilibrium for a $1-O(n^{-3\delta})$ fraction of agents to truthfully report their data;
            \item the private estimator $\bar{\theta}^P(\hat{D})$ is $\widetilde{O}(n^{-1+2\delta})$-accurate;
            \item it is individually rational for a $1-O(n^{-3\delta})$ fraction of agents to participate in the mechanism;
            \item the total expected budget required by the analyst is $\widetilde{O}(n^{-4\delta+1})$.
        \end{itemize}
    \end{corollary}
   \vspace{-0.2in}
    \subsubsection{Poisson Regression}
       \vspace{-0.1in}
    \begin{example}
        For a count-valued response $y\in \mathcal{Y}\equiv \{0,1,2,\cdots\}$, suppose its distribution is given by $p(y)=\frac{\lambda^y}{y!}e^{-\lambda}$ with parameter $\lambda>0$. If we set $\langle\mathbf{x}_i, \theta^{*}\rangle=\log \lambda$, then the distribution is equal to $\exp\{y\langle\mathbf{x}_i,\theta^{*}\rangle-
        \exp(\langle\mathbf{x}_i,\theta^{*}\rangle)-\log (y!)\}$. Thus, $A(a)=e^a$, $\phi=1$, and $c(y,\phi)=0$.
    \end{example}

    \begin{corollary}\label{cor:glm_poisson}
        %        Suppose that $F(\varepsilon, \gamma)\sim \varepsilon^4\gamma$ when $\varepsilon\to 0, \gamma\to 0$. 
        For any $\delta\in (\frac{1}{4}, \frac{1}{3})$ and $c>0$, we set $\bar{\mathcal{M}}=[n^{-\delta}, +\infty)$ 
        (we have $\Pi_{\mathcal{\bar{M}}}(\widetilde{y}_i)=\mathbf{1}_{\{\widetilde{y}_i= 0\}}n^{-\delta}+\mathbf{1}_{\{\widetilde{y}_i\neq 0\}}\widetilde{y}_i$), 
        $\varepsilon=n^{-3\delta}$, $\tau_2=\Theta(n^{\frac{1}{4}})$, $\alpha=\Theta(n^{-3\delta})$, $\beta=\Theta(n^{-c})$, $a_2=O(n^{-6\delta})$, $a_1=a_2(M_A+3M_A^2)+\tau_{\alpha,\beta}F(2\varepsilon, \gamma_n+2\gamma_{n/2})$. Then 
        the output of Mechanism~\ref{mech:GLM} satisfies $(O(n^{-3\delta}), O(n^{-\Omega(1)}))$-RJDP. And with probability at least $1-O(n^{-\Omega(1)})$, it holds that
        \begin{itemize}
            \item the symmetric threshold strategy $\sigma_{\tau_{\alpha,\beta}}$ is a $\widetilde{O}(n^{-4\delta})$-Bayesian Nash equilibrium for a $1-O(n^{-3\delta})$ fraction of agents to truthfully report their data;
            \item the private estimator $\bar{\theta}^P(\hat{D})$ is $\widetilde{O}(n^{-1+3\delta})$-accurate;
            \item it is individually rational for a $1-O(n^{-3\delta})$ fraction of agents to participate in the mechanism;
            \item the total expected budget required by the analyst is $\widetilde{O}(n^{-4\delta+1})$.
        \end{itemize}
    \end{corollary}
    \vspace{-0.3in}
    \section{Heavy-tailed Case for Linear Regression}
       \vspace{-0.1in}
    \label{sec:linear}
    In the previous section, we studied GLMs with sub-Gaussian covariates. However, the sub-Gaussian assumption is quite strong in practice. %The assumption that $\mathbf{x}_i$ are sampled from sub-Gaussian distribution in the above case may not hold in many complex systems.
    %For example, it is well known that macroeconomic variables and gene expressions in high dimensional datasets can have heavy or moderately heavy tails.
    For example, it has been widely known that  large-scale imaging datasets in biological studies and macroeconomic variables are corrupted by heavy-tailed noises due to limited measurement precision~\cite{fan2021shrinkage,biswas2007statistical}, which reveal that heavy-tailed distribution is a stylized feature of high-dimensional data. Thus, one natural question is whether we can extend the setting to the case where the data distribution is heavy-tailed. In this section, we focus on the linear regression and leave the generalized linear models as future work. Specifically, we consider the case where both covariate vectors $\mathbf{x}_i$ and responses $y_i$ only have bounded fourth moments.
    It is notable that Assumption~\ref{ass:6} is commonly used in the previous study in robust statistics~\cite{fan2021shrinkage,hu2022high}.
    \begin{assumption}\label{ass:6}
        We assume that $x_1, x_2, \cdots, x_n$ are i.i.d. and for each $x_i$ there exist constants $R_1, R_2=O(1)$ such that  $\sup_{\nu\in\mathcal{S}^{d-1}}\bb{E}(\nu^T\mathbf{x}_i)^4\leq  R_1$ for any unit vectors $\nu\in \mathbb{R}^d$ and $\bb{E}[y_i^4]\leq R_2$. Moreover, the covariance matrix $\Sigma$ of $\mathbf{x}_i$ satisfies that $\|\Sigma\|_\infty\geq \kappa_\infty$ and $\|\Sigma\|_2 \geq \kappa_2$ for some constants $\kappa_{\infty},\kappa_2=\Theta(1)$, {\em i.e.},  $\forall w\in \bb{R}^d$, $\|\Sigma w\|_{\infty}\geq \kappa_{\infty}\|w\|_{\infty}$ and $\|\Sigma w\|_2 \geq \kappa_2\|w\|_2$.  We also only focus on the low dimension case where $n=\tilde{\Omega}(d)$.
    \end{assumption}
    Before showing our method, we first discuss why $ \hat{\theta}(D)$ in~\eqref{est_glm} in the non-private case does not work in this setting. Recall that in the case of linear model, as shown in Section~\ref{subsec:linear} we have $ \hat{\theta}(D)=(X^TX)^{-1}X^T\tilde{y}$ in~\eqref{est_glm}. However, as now each $\mathbf{x}_i$ is heavy-tailed, the previous Lemma~\ref{lem:sensitivity_glm} on the $\ell_2$-norm sensitivity will not hold as the terms $X^TX$ and $X^T\tilde{y}$ will not be concentrated with high probability. %Due to the heavy-tailed corruption, performing the least squares method on the unprocessed data $(X, y)$ to approximate $\theta^{*}$ will lead to poor behavior (low accuracy and high sensitivity) of the estimator.
    Inspired by~\cite{fan2021shrinkage}, similar to $\tilde{y}_i$, here we also need to preprocess each $x_i$. Specifically,  we  apply a similar clipping operation as in the previous section to $y_i$ and an {\bf $l_4$-norm shrinkage} operation to $\mathbf{x}_i$, i.e., let $\widetilde{\mathbf{x}}_i$ satisfies  $\widetilde{\mathbf{x}}_i=\min\{\|\mathbf{x}_i\|_4, \tau_1\}\mathbf{x}_i/\|\mathbf{x}_i\|_4$ and $\widetilde{y}_i=\mathrm{sgn}(y_i)\min\{|y_i|, \tau_2\}$ for each $i\in [n]$, where $\tau_1, \tau_2>0$ are predetermined threshold values. Since now each $\tilde{x}_i$ and $\tilde{y_i}$ are  bounded, the terms of  $\tilde{X}^T\tilde{X}$ and $\tilde{X}^T\tilde{y}$ will be concentrated with high probability by Hoeffding’s inequality. In total, now the non-private estimator becomes to
    \begin{align*}
        \hat{\theta}(D)=(\widetilde{X}^T\widetilde{X})^{-1}\widetilde{X}^T\widetilde{y}.
        \numberthis\label{est_linear}
    \end{align*}
    Similar to the sub-Gaussian case we then project $\hat{\theta}^P(D)=\hat{\theta}(D)+\text{noise}$ onto the $\ell_2$-norm ball:
    $ \bar{\theta}^P(D)=\Pi_{\tau_\theta}(\hat{\theta}^P(D)).$ In the following we show the $\ell_2$-norm sensitivity and the accuracy of $\hat{\theta}(D)$.
    
    \begin{lemma}[Sensitivity]
        \label{lem:sensitivity_linear}
        If we set $\tau_1=\Theta((n/\log n)^{1/4})$ and $\tau_2=\Theta((n/\log n)^{1/8})$, then with probability at least $1-C_1 n^{-\Omega(1)}$,  the $\ell_2$-norm sensitivity  of $\hat{\theta}(D)$ computed from~\eqref{est_linear} satisfies
        \begin{align*}
            \max_{D\sim D'}\| \hat{\theta}(D)-\hat{\theta}(D')\|_2\leq   \Delta_{n}= C_0d^{\frac{3}{4}}(\frac{ \log n}{n})^{\frac{1}{8}},
        \end{align*}
        where $C_0, C_1>0$ are constants.
    \end{lemma}
    \begin{lemma}
        [Accuracy of the non-private estimator]
        \label{lem:accuracy_linear}
        By setting $\tau_1, \tau_2$ as in Lemma~\ref{lem:sensitivity_linear}, with probability at least $1-O(n^{-\Omega(1)})$, one has with some constant $C>0$: 
          $$\|\hat{\theta}(D)-\theta^{*}\|_2\leq Cd(\frac{\log n}{n})^{\frac{1}{4}}.$$
    \end{lemma}
    \begin{remark}
        Compared with the previous linear regression in the sub-Gaussian case, we can see due to the clipping parameters $\tau_1$ and $\tau_2$, both sensitivity and accuracy become larger. Secondly, besides the $\ell_4$-norm shrinkage operator in (\ref{est_linear}), another way is performing the element-wise shrinkage operator to each $x_i$ \cite{hu2022high},  i.e., $\widetilde{\mathbf{x}}_i$ satisfies  $\widetilde{x}_{ij}=\mathrm{sgn}(x_{ij})\min\{|x_{ij}|,\tau_1\}$ for each $i\in [n], j\in [d]$. However, by a similar proof as in Lemma \ref{lem:sensitivity_linear} one can see the $\ell_2$-norm sensitivity of $\hat{\theta}(D)$ based on this shrinkage operation will be $\tilde{O}(d^{\frac{3}{2}}(\frac{1}{n})^{\frac{1}{8}})$, which is larger than the bound in Lemma \ref{lem:sensitivity_linear}. Thirdly, while there are some work also uses different shrinkage operators (such as $\ell_2$-norm shrinkage or element-wise shrinkage) to variants statistical estimation problems \cite{hu2022high,fan2021shrinkage}, here we use the $\ell_4$-norm shrinkage and the thresholds $\tau_1, \tau_2$ are not equal, which are quite different with the previous approaches. 
    \end{remark}
    Based on the previous two lemmas and the similar idea as in the sub-Gaussian case, we propose Mechanism~\ref{mech:linear}. Specifically, we have the following results under Assumptions \ref{assump1:theta&lowD}-\ref{assump4:privacy_cost_coefficient_tail} and \ref{ass:6}.
    \begin{theorem}[Privacy]
        \label{thm:privacy_linear}
        Mechanism~\ref{mech:linear} satisfies $(2\varepsilon,\gamma_n+2\gamma_{n/2})$-random joint differential privacy, where $\gamma_n=C_1n^{-\Omega(1)}$ is the failure probability in Lemma  \ref{lem:sensitivity_linear}. 
    \end{theorem}
    
    \begin{theorem}[Truthfulness]
        \label{thm:truthfulness_linear}
         Fix a privacy parameter $\varepsilon$, a participation goal $1-\alpha$ and a desired confidence parameter $\beta$ in Definition~\ref{def:alpha_beta}. Then with probability at least $1-\beta-O(n^{-\Omega(1)})$, the symmetric threshold strategy $\sigma_{\tau_{\alpha,\beta}}$ is an $\eta$-approximate Bayesian Nash equilibrium in Mechanism~\ref{mech:linear} with
        \begin{align*}
            \eta=\widetilde{O}(a_2(\alpha^2d^2n^{\frac{9}{4}}+d^4n^{\frac{1}{4}}\varepsilon^{-2})
            +\tau_{\alpha,\beta}F(2\varepsilon, \gamma_n+2\gamma_{n/2})).
        \end{align*}
    \end{theorem}

    \begin{theorem}[Accuracy]
        \label{thm:accuracy_linear}
         Fix a privacy parameter $\varepsilon$, a participation goal $1-\alpha$ and a desired confidence parameter $\beta$ in Definition~\ref{def:alpha_beta}.  Then under the symmetric threshold strategy $\sigma_{\tau_{\alpha,\beta}}$, the output $\bar{\theta}^P(\hat{D})$ of Mechanism~\ref{mech:linear} satisfies that with probability at least $1-\beta-O(n^{-\Omega(1)})$,
        \begin{align*}
            \bb{E}\|\bar{\theta}^P(\hat{D})-\theta^{*}\|_2^2
            =\widetilde{O}(\alpha^2d^{\frac{3}{2}}n^{\frac{7}{4}}
            +d^{\frac{7}{2}}n^{-\frac{1}{4}}\varepsilon^{-2}
            ).
        \end{align*}
    \end{theorem}

    \begin{Mechanism}[htbp]
        \label{mech:linear}
        \caption{Private Heavy-tailed Linear Model Mechanism}
        Ask all agents to report their data $\hat{D}_1, \cdots, \hat{D}_n$\;
        Randomly partition agents into two groups, with respective data pairs $\hat{D}^0, \hat{D}^1$\;
        Compute estimators $\hat{\theta}(\hat{D}), \hat{\theta}(\hat{D}^0),  \hat{\theta}(\hat{D}^1)$ according to~\eqref{est_linear} with $\tau_1$ and $\tau_2$ in Lemma \ref{lem:sensitivity_linear}\;
        Compute  the sensitivity upper bounds $\Delta_n, \Delta_{n/2}$ in Lemma \ref{lem:sensitivity_linear}, and set differential privacy parameter $\varepsilon$\;
        Draw $v\in \bb{R}^d$ according to distribution $p(v)\propto \exp(-\frac{\varepsilon}{\Delta_n}\|v\|_2)$, and independently draw $v_{0},v_{1}\in \bb{R}^d$ according to distribution
        $p(v)\propto \exp(-\frac{\varepsilon}{\Delta_{n/2}}\|v\|_2)$\;
        Add noise: $\hat{\theta}^P(\hat{D})=\hat{\theta}(\hat{D})+v, \hat{\theta}^{P}(\hat{D}^b)=\hat{\theta}({\hat{D}^b})+v_{b}$ for $b=0,1$\;
        Compute private estimators $\bar{\theta}^P(\hat{D})=\Pi_{\tau_\theta}(\hat{\theta}^P(\hat{D}))$ and $\bar{\theta}^{P}(\hat{D}^b)=\Pi_{\tau_\theta}(\hat{\theta}^{P}(\hat{D}^b))$ for $b=0,1$\;
        Set parameters $a_1, a_2$, and compute payments to each agent $i$:  if agent $i$'s is in group $1-b$, then he will receive payment
        \begin{align*}
            \pi_i=B_{a_1,a_2}\left(\langle \mathbf{\widetilde{x}}_i,\bar{\theta}^{P}(\hat{D}^{b})\rangle,\langle \mathbf{\widetilde{x}}_i, \bb{E}_{\theta\sim p(\theta|\hat{D}_i)}[\theta] \rangle\right).
        \end{align*}
    \end{Mechanism}
    \begin{theorem}[Individual rationality]
        \label{thm:rationality_linear}
        With probability at least $1-\beta-O(n^{-\Omega(1)})$, Mechanism~\ref{mech:linear} is individually rational for all agents with cost coefficients $c_i\leq \tau_{\alpha,\beta}$ as long as
        \begin{align*}
            a_1\geq a_2(d^{\frac{1}{4}}\tau_1\tau_{\theta}+
            3d^{\frac{1}{2}}\tau_1^2\tau_{\theta}^2)+\tau_{\alpha,\beta}F(2\varepsilon, \gamma_n+2\gamma_{n/2}), 
        \end{align*}
        regardless of the reports from agents with cost coefficients above $\tau_{\alpha,\beta}$.
    \end{theorem}

    \begin{theorem}[Budget]
        \label{thm:budget_linear}
        The total expected budget $\mathcal{B}:=\bb{E}[\sum_{i=1}^{n}\pi_i]$ required by the analyst to run Mechanism \ref{mech:linear} under threshold equilibrium strategy $\sigma_{\tau_{\alpha,\beta}}$ satisfies 
        \begin{align*}
            \mathcal{B}=\widetilde{O}(n(a_2\sqrt{dn}+\tau_{\alpha,\beta}F(2\varepsilon, \gamma_n+2\gamma_{n/2}))).
        \end{align*}
    \end{theorem}
    
    \begin{corollary}\label{cor:heavy_linear}
        Suppose that $F(\varepsilon, \gamma)=(1+\gamma)\varepsilon^9$. For any $\delta\in (\frac{1}{9}, \frac{1}{8})$ and $c>0$, we set $\bar{\mathcal{M}}=\mathbb{R}$, $\tau_1=\Theta((n/\log n)^{1/4})$, $\tau_2=\Theta((n/\log n)^{1/8})$, $\varepsilon=n^{-\delta}$, $\alpha=\Theta(n^{-1+\delta})$, $\beta=\Theta(n^{-c})$, $a_2=n^{-\frac{1}{2}-9\delta}$, $a_1=a_2(d^{\frac{1}{4}}\tau_1\tau_{\theta}+3d^{\frac{1}{2}}\tau_1^2\tau_{\theta}^2)+\tau_{\alpha, \beta}F(2\varepsilon, \gamma_n+2\gamma_{n/2})$. Then 
        the output of Mechanism~\ref{mech:linear} satisfies $(O(n^{-\delta}), O(n^{-\Omega(1)}))$-random joint differential privacy. And with probability at least $1-O(n^{-\Omega(1)})$, it holds that
        \begin{itemize}
            \item the symmetric threshold strategy $\sigma_{\tau_{\alpha,\beta}}$ is a $\widetilde{O}(n^{-9\delta})$-Bayesian Nash equilibrium for a $1-O(n^{-1+\delta})$ fraction of agents to truthfully report their data;
            \item the private estimator $\bar{\theta}^P(\hat{D})$ is $\widetilde{O}(n^{-\frac{1}{4}+2\delta})$-accurate;
            \item it is individually rational for a $1-O(n^{-1+\delta})$ fraction of agents to participate in the mechanism;
            \item the total expected budget required by the analyst is $\widetilde{O}(n^{-9\delta+1})$.
        \end{itemize}
    \end{corollary}
    %    1. Summarize the main contributions of the paper, the results, the conclusions in one paragraph.
    %    2. Bring up some (obvious) limitations of the assumptions. Describe why you have bounded the impact of these assumptions in your results.
    %    3. Bring up some (obvious) future steps for this approach, and motivate them in terms of the new domain-specific results from these future steps.
    %    Note: Don't dig deep here - these obvious ones will be most helpful for readers getting to know this work, otherwise you will lose the readers and appear as if you overlooked the obvious limitations of your approach.
    
    %    \section*{References}
    %    \label{references}
    
    %    Borrowing any more than about three words from a paper means you put that phrase in quotes and cite the paper from which it came.
    %    Cite often and freely. Giving respect to related work, or thoughtful work for that matter, never goes unappreciated.

\begin{remark}
    It is worth mentioning that throughout the whole paper, the privacy cost function $f_i(c_i, \epsilon, \gamma)$ can change with respect to the distortion in the report, i.e. $|y_i-\hat{y}_i|$. But we do not write explicitly this relation, since it does not matter in reaching our results above. What really matters is the assumption on upper bounding the privacy cost function (see Assumption~\ref{assump2:privacy_cost_function_bound}). And we only consider the case where an agent can reduce his/her privacy cost \textit{at most} by misreporting the data (see the last paragraph in the proof of Theorem~\ref{thm:truthfulness_glm}). However, the lack of a closer look at the distribution of an agent’s utility under different reporting strategies prevents us concluding that each agent will have the highest utility \textit{in average} if she truthfully reports the data  (We only establish that the symmetric threshold strategy is an $\eta$ Bayesian Nash equilibrium, which means an agent may increase $\eta$ utility by misreporting the data). Thus, we need to further study how the payment and privacy cost vary under different reporting strategies to obtain a more satisfactory result. 
\end{remark}

\section*{Acknowledgement}
Di Wang is supported in part by the baseline funding BAS/1/1689-01-01, funding from the CRG grand URF/1/4663-01-01, FCC/1/1976-49-01 from CBRC and funding from the AI Initiative REI/1/4811-10-01 of King Abdullah University of Science and Technology (KAUST).  He is also supported by the funding of the SDAIA-KAUST Center of Excellence in Data Science and Artificial Intelligence (SDAIA-KAUST AI). Jinyan Liu is partially supported by National Natural Science Foundation of China (NSFC Grant No.62102026). 
    
    \bibliographystyle{plain}
    \bibliography{references}
    
    \newpage
    \appendix
    
    \section{Supporting lemmas}
    \label{app:supporting lemmas}
    \begin{lemma}\label{lem:proj}
        For any $w, w'\in \mathbb{R}^d$ and closed convex set  $\mathcal{C}\subseteq \mathbb{R}^d$ we have
        \begin{equation*}
            \|\Pi_\mathcal{C}(w)-\Pi_\mathcal{C}(w')\|_2\leq \|w-w'\|_2,
        \end{equation*}
        where $\Pi_\mathcal{C}$ is the projection operation onto the set $\mathcal{C}$, i.e., $\Pi_\mathcal{C}(v)=\arg\min_{u\in \mathcal{C}}\|u-v\|_2$.
    \end{lemma}
    \begin{lemma}[\cite{vershynin2018high}]
        \label{lem:subG}
        Let $X_1,\cdots, X_n$ be $n$ independent random variables such that $X_i\sim \mathrm{subG}(\sigma^2)$. Then for any $a\in \bb{R}^{n}, t>0$, we have
        \begin{align*}
            \bb{P}(|\sum_{i=1}^{n}a_iX_i|>t)\leq 2\exp(-\frac{t^2}{2\sigma^2 \|a\|_2^2}).
        \end{align*}
    \end{lemma}
    
    \begin{lemma}[Hoeffding's inequality \cite{vershynin2018high}]
        \label{lem:Hoeffding}
        Let $X_1,\cdots, X_n$ be independent random variables bounded by the interval $[a,b]$. Then, for any $t>0$,
        \begin{align*}
            \bb{P}(|\frac{1}{n}\sum_{i=1}^{n}X_i-\frac{1}{n}\sum_{i=1}^{n}\bb{E}[X_i]|>t)\leq 2\exp(-\frac{2nt^2}{(b-a)^2}).
        \end{align*}
    \end{lemma}
    
     \begin{lemma}[Bernstein's inequality \cite{vershynin2018high}]
        \label{lem:Bernstein_ineq}
        Let $X_1, X_2, \cdots, X_n$ be independent centered bounded random variables, i.e. $|X_i|\leq M$ and $\bb{E}[X_i]=0$, with variance $\bb{E}[X_i^2]=\sigma^2$. Then, for any $t>0$,
        \begin{align*}
            \bb{P}(|\sum_{i=1}^{n}X_i|>\sqrt{2n\sigma^2t}+\frac{2Mt}{3})\leq 2e^{-t}.
        \end{align*}
    \end{lemma}
    
    \begin{lemma}[Billboard lemma~\cite{hsu2016private}]
        \label{lem:billboard}
        Let $\mathcal{M}:\mathcal{D}^n\to \mathcal{O}$ be an $\varepsilon$-differential private mechanism. Consider a set of $n$ functions $\pi_i:\mathcal{D}\times \mathcal{O}\to \mathcal{R}$, for $i\in [n]$. Then the mechanism $\mathcal{M}^{\prime}:\mathcal{D}^n\to  \mathcal{O}\times\mathcal{R}^n$ that computes $r=\mathcal{M}(D)$ and outputs $\mathcal{M}^{\prime}(D)=(r,\pi_1(D_1,r),\cdots ,\pi_n(D_n,r))$, where $D_i$ is the agent $i$'s data, is $\varepsilon$-differential private.
    \end{lemma}
    
    \begin{lemma}
        \label{lem:moment}
        If $v\in \bb{R}^d$ is drawn from the distribution with probability density function $p(v)\propto \exp(-\frac{\varepsilon}{\Delta}\|v\|_2)$, then
        $\bb{E}[v]=0$, $\bb{E}[\|v\|_2^2]=d(d+1)(\frac{\Delta}{\varepsilon})^2$, $\bb{E}[\|v\|_2]=\frac{d\Delta}{\varepsilon}$.
    \end{lemma}
    
    \begin{lemma}
        \label{lem:sensitivity_k_entries}
        Let $\hat{\theta}(D)$ and $\hat{\theta}(D^{\prime})$ be the estimators on two fixed datasets $D, D^{\prime} $ that differ on at most $k$ entries. Suppose that with probability at least $1-\gamma_n$, the sensitivity of $\hat{\theta}(D)$ is upper bounded by $\Delta_{n}$. Then we have
        with probability at least $1-k\gamma_n$, it holds that
        \begin{align*}
            \|\hat{\theta}(D)-\hat{\theta}(D^{\prime})\|_2\leq k\Delta_n.
        \end{align*}
    \end{lemma}
    
    \begin{lemma}[Bound on threshold $\tau_{\alpha, \beta}$]
        \label{lem:bound_on_threshold}
        Under the Assumption~\ref{assump4:privacy_cost_coefficient_tail}, $\tau_{\alpha,\beta}\leq \frac{1}{\lambda}\log\frac{1}{\alpha\beta}$.
    \end{lemma}
    
    \begin{lemma}[Largest singular value of sub-Gaussian matrices~\cite{vershynin2010introduction}]
        \label{lem:singular_value}
        Let $A$ be an $n\times d$ matrix whose rows $A_i$ are independent sub-Gaussian isotropic (i.e. $\bb{E}[A_iA_i^T]=I$) random vectors in $\bb{R}^d$. Then for every $t\geq 0$, with probability at least $1-2e^{-c_0t^2}$ one has
        \begin{align*}
            \|A\|_2\leq \sqrt{n}+C_0\sqrt{d}+t.
        \end{align*}
        where $c_0, C_0\geq 0$ are universal constants.
    \end{lemma}
    
    \begin{lemma}[Covariance estimation for sub-Gaussian distribution~\cite{vershynin2010introduction}]
        \label{lem:covariance_estimation}
        Assume that $X$ be an $n\times d$ matrix whose rows $\mathbf{x}_i^T$ are independent sub-Gaussian random vectors in $\bb{R}^n$ with covariance matrix $\Sigma$. Then for every $s\geq 0$, with probability at least $1-2\exp(-c_1s^2)$, one has
        \begin{align*}
            \|\frac{X^TX}{n}-\Sigma\|_2\leq \max(\delta,\delta^2)\quad\text{where}\quad \delta=C_1\sqrt{\frac{d}{n}}+\frac{s}{\sqrt{n}},
        \end{align*}
        where $c_1,C_1>0$ are universal constants.
    \end{lemma}
    
    \begin{lemma}[Covariance estimation for heavy-tailed distribution]
        \label{lem:cov_est_heavytailed}
        Assume that $X$ be an $n\times d$ matrix whose rows $\mathbf{x}_i^T$ are independent random vectors in $\bb{R}^d$ with finite fourth moment, i.e., $\sup_{\nu\in\mathcal{S}^{d-1}}\bb{E}(\nu^T\mathbf{x}_i)^4\leq R<\infty$, where $\mathcal{S}^{d-1}$ is a $d$ dimensional unit sphere. Denote the covariance matrix as $\Sigma:=\bb{E}[\mathbf{x}_i\mathbf{x}_i^T]$. For any $\delta>0$, let $\tau_1=\Theta((nR/(\delta\log n))^{1/4})$ and  $\widetilde{\mathbf{x}}_i=\min\{\|\mathbf{x}_i\|_4, \tau_1\}\mathbf{x}_i/\|\mathbf{x}_i\|_4$ for each $i\in [n]$. Then, with probability at least $1-dn^{-C\delta}$ it holds that
        \begin{align*}
            \|\frac{\widetilde{X}^T\widetilde{X}}{n}-\Sigma\|_{2}\leq
            2\sqrt{\frac{\delta Rd\log n}{n}},
        \end{align*}
        where constant $C>0$ is a universal constant.
    \end{lemma}
    
    \section{Proofs of Supporting Lemmas}
    \label{app:proofs}
    \begin{proof}[{\bf Proof of Lemma \ref{lem:proj}}]
Denote $b= \Pi_\mathcal{C}(w)$ and $b'= \Pi_\mathcal{C}(w')$. Since $b$ and $b'$ are in $\mathcal{C}$, so the segment $bb'$ is contained in $\mathcal{C}$, thus we have for all $t\in [0, 1], \|(1-t)b+tb'-w\|_2\geq \|b-w\|_2$. Thus 
\begin{equation*}
    0 \leq  \frac{d}{dt}\|tb+(1-t)b'-w\|_2^2|_{t=0}=2\langle b'-b, b-w\rangle 
\end{equation*}
Similarly, we have $\langle b-b', b'-w' \rangle\geq 0$. Now consider the function $D(t)= \|(1-t)b+tw- (1-t)b'-tw'\|_2^2=\|b-b'+t(w-w'+b'-b)\|_2^2$, which is a quadratic function in $t$. And by the previous two inequalities  we have $D'(0)=2 \langle b-b',  w-w'+b'-b\rangle\geq 0$. Thus $D(\cdot)$ is a increasing function on $[0, \infty)$, thus $D(1)\geq D(0)$ which means $\|w-w'\|_2\geq \|b-b'\|_2$. 
\end{proof}

    \begin{proof}[ {\bf Proof of Lemma~\ref{lem:moment}}]
        Write $p(v)=\frac{1}{Z}\exp(-\frac{\varepsilon}{\Delta}\|v\|_2)$, in which $Z$ is a constant such that $\int_{\bb{R}^d}p(v)\dd v=1$. Then
        \begin{align*}
            Z = \int_{\bb{R}^d}\exp(-\frac{\varepsilon}{\Delta}\|v\|_2)\dd v =\int_{0}^{\infty}\exp(-\frac{\varepsilon}{\Delta}r)A_dr^{d-1}\dd r
            = A_d(d-1)!(\frac{\Delta}{\varepsilon})^d,
        \end{align*}
        where $A_d$ is the "area" of boundary of $d$-dimensional unit ball and the last inequality follows from integration by parts for $d-1$ times. Similarly,
        \begin{align*}
            \bb{E}[\|v\|_2^2]
            =\int_{\bb{R}^d}\frac{1}{Z}\exp(-\frac{\varepsilon}{\Delta}\|v\|_2)\|v\|_2^2\dd v
            &=\int_{0}^{\infty}\frac{1}{Z}\exp(-\frac{\varepsilon}{\Delta}r)A_dr^{d+1}\dd r\\
            &=\frac{1}{Z}A_d(d+1)!(\frac{\Delta}{\varepsilon})^{d+2}
            =d(d+1)(\frac{\Delta}{\varepsilon})^2,
        \end{align*}
        and $\bb{E}[\|v\|]=\frac{d\Delta}{\varepsilon}$.  Since $p(v)$ is symmetric to the origin, $\bb{E}[v]=0$.
    \end{proof}

    \begin{proof}[{\bf Proof of Lemma~\ref{lem:sensitivity_k_entries}}]
        Define a sequence of datasets $D^0,D^1,\cdots,D^k$, such that $D^0=D$, $D^k=D^{\prime}$, and for each $i\in [k]$, $D^i, D^{i-1}$ differ on at most one agent's dataset. Then, by the triangular inequality, we obtain
        \begin{align*}
            \|\hat{\theta}(D)-\hat{\theta}(D^{\prime})\|_2=
            \|\hat{\theta}(D^0)-\hat{\theta}(D^k)\|_2=
            \|\sum_{i=1}^{k}\hat{\theta}(D^{i-1})-\hat{\theta}(D^i)\|_2\leq
            \sum_{i=1}^{k}\|\hat{\theta}(D^{i-1})-\hat{\theta}(D^i)\|_2 \leq k\Delta_n.
        \end{align*}
        with probability at least $1-k\gamma_n$ by taking a union bound over $k$ failure probabilities $\gamma_n$.
    \end{proof}

    \begin{proof}[{\bf Proof of Lemma~\ref{lem:bound_on_threshold}}]
        We first bound $\tau_{\alpha,\beta}^1$. Since $n=\#\{i:c_i\leq \tau\}+\#\{i:c_i>\tau\}$, the event  $\{\#\{i:c_i\leq \tau\}\geq (1-\alpha)n\} $ is equivalent to the event $\{\#\{i:c_i> \tau\}\leq  \alpha n\}$. Thus, by the definition of $\tau_{\alpha,\beta}^1$,
        \begin{align*}
            \tau_{\alpha,\beta}^1
            &=\inf\{\tau>0: \bb{P}_{(c_1,\cdots,c_n)\sim p^{n}}(\#\{i:c_i> \tau\}\leq  \alpha n)\geq 1-\beta\}\\
            & =\inf\{\tau>0: \bb{P}_{(c_1,\cdots,c_n)\sim p^{n}}(\#\{i:c_i> \tau\}>\alpha n)\leq \beta\},
        \end{align*}
        By Markov's inequality, we have
        \begin{align*}
            &\bb{P}_{(c_1,\cdots,c_n)\sim p^{n}}(\#\{i:c_i> \tau\}>\alpha n)\leq \frac{\bb{E}_{(c_1,\cdots,c_n)\sim p^{n}}[\sum_{i=1}^{n}\mathbf{1}_{\{c_i>\tau\}}]}{\alpha n}\\
            &=\frac{\sum_{i=1}^{n}\bb{E}_{c_i\sim p}[\mathbf{1}_{\{c_i>\tau\}}]}{\alpha n}=\frac{n\bb{P}[c_i>\tau]}{\alpha n}=\frac{\bb{P}[c_i>\tau]}{\alpha}.
        \end{align*}
        Thus, $\{\tau>0: \bb{P}(c_i>\tau)\leq \alpha \beta\}\subseteq \{\tau>0: \bb{P}_{(c_1,\cdots,c_n)\sim p^{n}}(\#\{i:c_i> \tau\}>\alpha n)\leq \beta\}$, which implies $\tau_{\alpha,\beta}^1\leq \inf\{\tau>0: \bb{P}(c_i>\tau)\leq \alpha \beta\}$. The Assumption~\ref{assump4:privacy_cost_coefficient_tail} implies that  $\bb{P}(c_i> \tau)\leq e^{-\lambda\tau}$. Hence, $\tau_{\alpha,\beta}^1\leq \frac{1}{\lambda}\log \frac{1}{\alpha\beta}$. By the definition of $\tau_{\alpha}^2$ and Assumption~\ref{assump4:privacy_cost_coefficient_tail},  we have $\tau_{\alpha}^2\leq \frac{1}{\lambda}\ln \frac{1}{\alpha}$. Since $\beta\in(0,1)$, $\frac{1}{\lambda}\log \frac{1}{\alpha\beta}>\frac{1}{\lambda}\log\frac{1}{\alpha}$, then $\tau_{\alpha,\beta}=\max\{\tau_{\alpha,\beta}^1, \tau_{\alpha}^2\}\leq \frac{1}{\lambda}\log \frac{1}{\alpha\beta}$.
    \end{proof}
    
    \begin{proof}[{\bf Proof of Lemma~\ref{lem:cov_est_heavytailed}}]
        Note that
        \begin{align*}
            \|\widetilde{\mathbf{x}}_i\widetilde{\mathbf{x}}_i^T\|_2
            =\sup_{v\in\mathcal{S}^{n-1}}|\nu^T\widetilde{\mathbf{x}}_i\widetilde{\mathbf{x}}_i^T\nu|
            =\sup_{\nu\in\mathcal{S}^{d-1}}|\nu^T\widetilde{\mathbf{x}}_i|^2
            =\|\widetilde{\mathbf{x}}_i\|_2^2\leq \sqrt{d}\|\widetilde{\mathbf{x}}_i\|_4^2\leq \sqrt{d}\tau^2,
        \end{align*}
        and
        \begin{align*}
            \|\bb{E}\widetilde{\mathbf{x}}_i\widetilde{\mathbf{x}}_i^T\|_2
            =\sup_{\nu\in\mathcal{S}^{d-1}}|\nu^T\bb{E}\widetilde{\mathbf{x}}_i\widetilde{\mathbf{x}}_i^T\nu|
            =\sup_{\nu\in\mathcal{S}^{d-1}}\bb{E}|\nu^T\widetilde{\mathbf{x}}_i|^2
            \leq \sup_{\nu\in\mathcal{S}^{d-1}}\bb{E}|\nu^T\mathbf{x}_i|^2
            \leq \sup_{\nu\in\mathcal{S}^{d-1}}\sqrt{\bb{E}(\nu^T\mathbf{x}_i)^4}=\sqrt{R}.
        \end{align*}
        Thus,
        \begin{align*}
            \|\widetilde{\mathbf{x}}_i\widetilde{\mathbf{x}}_i^T-\bb{E}\widetilde{\mathbf{x}}_i\widetilde{\mathbf{x}}_i^T\|_2
            \leq \|\widetilde{\mathbf{x}}_i\widetilde{\mathbf{x}}_i^T\|_2+
            \|\bb{E}\widetilde{\mathbf{x}}_i\widetilde{\mathbf{x}}_i^T\|_2
            \leq \sqrt{d}\tau^2+\sqrt{R}.
        \end{align*}
        Also note that
        \begin{align*}
            \|\bb{E}(\widetilde{\mathbf{x}}_i\widetilde{\mathbf{x}}_i^T)^2\|_2
            &=\sup_{\nu\in\mathcal{S}^{d-1}}|\nu^T\bb{E}(\widetilde{\mathbf{x}}_i\widetilde{\mathbf{x}}_i^T)^2\nu|
            =\sup_{\nu\in\mathcal{S}^{d-1}}\bb{E}(\nu^T\widetilde{\mathbf{x}}_i)^2\|\widetilde{\mathbf{x}}_i\|_2^2
            \leq \sup_{\nu\in\mathcal{S}^{d-1}}\sum_{i=j}^{d}\bb{E}x_{ij}^2(\nu^T\mathbf{x}_i)^2\\
            &\leq \sup_{\nu\in\mathcal{S}^{d-1}}\sum_{j=1}^{d}\sqrt{\bb{E}(x_{ij}^4)\bb{E}(\nu^T\mathbf{x}_i)^4}\leq Rd.
        \end{align*}
        Thus,
        \begin{align*}
            \|\bb{E}(\widetilde{\mathbf{x}}_i\widetilde{\mathbf{x}}_i^T-\bb{E}\widetilde{\mathbf{x}}_i\widetilde{\mathbf{x}}_i^T)^2\|_2
            =\|\bb{E}(\widetilde{\mathbf{x}}_i\widetilde{\mathbf{x}}_i^T)^2-(\bb{E}\widetilde{\mathbf{x}}_i\widetilde{\mathbf{x}}_i^T)^2\|_2
            \leq \|\bb{E}(\widetilde{\mathbf{x}}_i\widetilde{\mathbf{x}}_i^T)^2\|_2
            +\|\bb{E}\widetilde{\mathbf{x}}_i\widetilde{\mathbf{x}}_i^T\|_2^2\leq R(d+1).
        \end{align*}
        By Theorem 5.29 in~\cite{vershynin2010introduction}, for any $t>0$, it holds that
        \begin{align*}
            \bb{P}(\|\frac{1}{n}\sum_{i=1}^{n}\widetilde{\mathbf{x}}_i\widetilde{\mathbf{x}}_i^T-\bb{E}\widetilde{\mathbf{x}}_i\widetilde{\mathbf{x}}_i^T\|_2>t)
            \leq 2d\exp\big[-c\min\{\frac{nt^2}{R(d+1)}, \frac{nt}{\sqrt{d}\tau^2+\sqrt{R}}\}\big].
            \numberthis\label{lem:cov_est_heavytailed1}
        \end{align*}
        where $c>0$ is a constant. In addition,
        \begin{align*}
            \|\bb{E}\widetilde{\mathbf{x}}_i\widetilde{\mathbf{x}}_i^T-\bb{E}\mathbf{x}_i\mathbf{x}_i^T\|_2
            &=\sup_{\nu\in\mathcal{S}^{d-1}}|\nu^T(\bb{E}\widetilde{\mathbf{x}}_i\widetilde{\mathbf{x}}_i^T-\bb{E}\mathbf{x}_i\mathbf{x}_i^T)\nu|\\
            &=\sup_{\nu\in\mathcal{S}^{d-1}}|\bb{E}(((\nu^T\widetilde{\mathbf{x}}_i)^2
            -(\nu^T\mathbf{x}_i)^2)\mathbf{1}_{\{\|\mathbf{x}_i\|_4>\tau\}})|\\
            &\leq\sup_{\nu\in\mathcal{S}^{d-1}}|\bb{E}(\nu^T\mathbf{x}_i)^2\mathbf{1}_{\{\|\mathbf{x}_i\|_4>\tau\}}|\\
            &\leq \sup_{\nu\in\mathcal{S}^{d-1}}\sqrt{\bb{E}(\nu^T\mathbf{x}_i)^4\bb{P}(\|\mathbf{x}_i\|_4>\tau)}\\
            &\leq \sup_{\nu\in\mathcal{S}^{d-1}}\sqrt{\bb{E}(\nu^T\mathbf{x}_i)^4\frac{\bb{E}\|\mathbf{x}_i\|_4^4}{\tau^4}}\leq \frac{R\sqrt{d}}{\tau^2}.
            \numberthis\label{lem:cov_est_heavytailed2}
        \end{align*}
        Let $\tau=\Theta((nR/(\delta\log n))^{1/4})$ and $t=\sqrt{\delta Rd\log n/n}$. Then, combining~\eqref{lem:cov_est_heavytailed1} and~\eqref{lem:cov_est_heavytailed2} delivers that with probability at least $1-2dn^{-C\delta}$ one has
        \begin{align*}
            \|\frac{\widetilde{X}^T\widetilde{X}}{n}-\Sigma\|_2
            \leq\|\frac{1}{n}\sum_{i=1}^{n}\widetilde{\mathbf{x}}_i\widetilde{\mathbf{x}}_i^T-\bb{E}\widetilde{\mathbf{x}}_i\widetilde{\mathbf{x}}_i^T\|_2
            +\|\bb{E}\widetilde{\mathbf{x}}_i\widetilde{\mathbf{x}}_i^T-\bb{E}\mathbf{x}_i\mathbf{x}_i^T\|_2\leq 2\sqrt{\frac{\delta Rd\log n}{n}}.
        \end{align*}
    \end{proof}

    \section{Omitted Proofs}
    \begin{proof}[{\bf Proof of Lemma \ref{lem:sensitivity_glm}}]
        Let $D$ and $D^{\prime}$ be two arbitrary neighboring datasets that differ only on the last agent's report $(\mathbf{x}_n, \hat{y}_n)$ in $D$ and $(\mathbf{x}_n^{\prime},\hat{y}_n^{\prime})$ in $D^{\prime}$. Let $\mathcal{E}:=\{\max_{i\in [n]}\|\mathbf{x}_i\|_2\leq 4\sqrt{c}\sigma\sqrt{\log n}\}$ for constant $c>1$. By Lemma~\ref{lem:subG}, we have $\bb{P}(\mathcal{E}^c)\leq n^{-{c+1}}$. In the following, we will always assume the event $\mathcal{E}$ holds.
        
        \noindent \textbf{Step 1:} Upper bound $\|\frac{X^T(A^{\prime})^{-1}(\proj(y))}{n}\|_2$. By Lemma~\ref{lem:singular_value}, and note that $\mathbf{x}_i^T\Sigma^{-\frac{1}{2}}$ is isotropic, then
        with probability at least $1-2e^{-c_0n}$ we have $\|X^T\|_2=\|X\|_2=\|X\Sigma^{-\frac{1}{2}}\Sigma^{\frac{1}{2}}\|_2\leq \|X\Sigma^{-\frac{1}{2}}\|_2\|\Sigma^{\frac{1}{2}}\|_2
        \leq (2\sqrt{n}+C_0\sqrt{d})\sqrt{\lambda_{\max}}$, where $\lambda_{\max}$ is the largest eigenvalue of $\Sigma$. Thus,
        \begin{align*}
            \|\frac{X^T(A^{\prime})^{-1}(\proj(\widetilde{y}))}{n}\|_2
            &\leq \frac{1}{n}\|X^T\|_2\|(A^{\prime})^{-1}(\proj(\widetilde{y}))\|_2\\
            &\leq
            \frac{1}{n}(2\sqrt{n}+C_0\sqrt{d})\sqrt{\lambda_{\max}}\sqrt{n}\kappa_{A,1}=O(\sqrt{\lambda_{\max}}\kappa_{A,1}).
        \end{align*}
        
        \noindent \textbf{Step 2:} Upper bound the  sensitivity of
        $\|\frac{X^T(A^{\prime})^{-1}(\proj(\widetilde{y}))}{n}\|_2$.
        \begin{align*}
            & \quad \|\frac{X^T(A^{\prime})^{-1}(\proj(\widetilde{y}))}{n}-
            \frac{X^{\prime^T}(A^{\prime})^{-1}(\proj(\widetilde{y}^{\prime}))}{n}\|_2  = \frac{1}{n} \|\mathbf{x}_n(A^{\prime})^{-1}(\proj(\widetilde{y}_n)-\mathbf{x}_n^{\prime}(A^{\prime})^{-1}(\proj(\widetilde{y}_n^{\prime}))\|_2 \\
            &\leq \frac{1}{n}\left(\|\mathbf{x}_n\|_2 |(A^{\prime})^{-1}(\proj(\widetilde{y}_n))|+
            \|\mathbf{x}_n^{\prime}\|_2 |(A^{\prime})^{-1}(\proj(\widetilde{y}_n^{\prime}))| \right)= O(\frac{1}{n}\sigma\sqrt{\log n}\kappa_{A,1}).
        \end{align*}
        
        \noindent \textbf{Step 3: }Upper bound  $\|(\frac{X^TX}{n})^{-1}\|_2$. For any nonzero vector $w\in \bb{R}^d$,
        Note that
        \begin{align*}
            \|\frac{X^TX}{n}w\|_2
            &=\|\frac{X^TX}{n}w-\Sigma w+\Sigma w\|_2\\
            &\geq \|\Sigma w\|_2-\|(\frac{X^TX}{n}-\Sigma)w\|_2
            \geq (\kappa_{2}-\|\frac{X^TX}{n}-\Sigma\|_2\|w\|_2).
            \numberthis\label{lem:sensitivity_lowD1}
        \end{align*}
        By Lemma~\ref{lem:covariance_estimation}, let $s=C_1t\sqrt{d}$ for any $t\geq 1$,  then when $n\geq 4C_1^2t^2d=\Omega(t^2d)$,
        with probability at least $1-2e^{-t^2d}$, we have $\|\frac{X^TX}{n}-\Sigma\|_2\leq 2C_1t\sqrt{\frac{d}{n}} $. Thus, when $n\geq \frac{16C_1^2t^2d}{\kappa_{2}^2}$, we have $\|\frac{X^TX}{n}-\Sigma\|_{2}\leq \frac{\kappa_{2}}{2}$. Combining this inequality and~\eqref{lem:sensitivity_lowD1} yields that $\|\frac{X^TX}{n}w\|_{2}\geq \frac{\kappa_{2}}{2}\|w\|_{2}$, which implies $\|(\frac{X^TX}{n})^{-1}\|_{2}\leq \frac{2}{\kappa_{2}}$.
        
        \noindent \textbf{Step 4: } Next we bound the sensitivity of $\|(\frac{X^TX}{n})^{-1}\|_2$. Note that for any two nonsingular square matrices $A, B$ with the same size, it holds that
        $A^{-1}-B^{-1}=-B^{-1}(A-B)A^{-1}$. Thus,
        \begin{align*}
            \|(\frac{X^TX}{n})^{-1}-(\frac{X^{\prime^T}X^{\prime}}{n})^{-1}\|_2
            &\leq \|(\frac{X^TX}{n})^{-1}\|_2\|(\frac{X^{\prime^T}X^{\prime}}{n})\|_2
            \|\frac{X^TX}{n}-\frac{X^TX}{n}\|_2\\
            &\leq \frac{4}{\kappa_{2}^2}(\|\frac{X^TX}{n}-\Sigma\|_2+\|\Sigma-\frac{X^{\prime^T}X^{\prime}}{n}\|_2)\leq \frac{16C_1t}{\kappa_{2}^2}\sqrt{\frac{d}{n}}.
        \end{align*}
        Take $t=\sqrt{\log n}$ we have $  \|(\frac{X^TX}{n})^{-1}-(\frac{X^{\prime^T}X^{\prime}}{n})^{-1}\|_2= O(\frac{1}{\kappa_2^2}\sqrt{\frac{d\log n}{n}})$.
        
        \noindent\textbf{ Step 5: }   Applying the inequality $\|AB-A^{\prime}B^{\prime}\|_2= \|AB-AB^{\prime}+AB^{\prime}-A^{\prime}B^{\prime}\|_2 \leq \|A\|_2\|B-B^{\prime}\|_2+\|A-A^{\prime}\|_2\|B^{\prime}\|_2$, we have with probability at least $1-n^{-c+1}-4n^{-d}$,
        \begin{align*}
            & \|\hat{\theta}(D)-\hat{\theta}(D')\|_2\\
            &=\|(\frac{X^TX}{n})^{-1}\frac{X^T(A^{\prime})^{-1}(\proj(\widetilde{y}))}{n}-(\frac{X^{\prime^T}X^{\prime}}{n})^{-1}\frac{X^{\prime^T}(A^{\prime})^{-1}(\proj(\widetilde{y}^{\prime}))}{n}\|_2\\
            &= O(\kappa_{A,1}(\frac{2\sigma}{\kappa_{2}}\frac{\sqrt{\log n}}{n}
            +\frac{\sqrt{\lambda_{\max}}}{\kappa_{2}^2}\sqrt{\frac{d\log n}{n}}),
            \numberthis\label{lem:sensitivity_lowD2}
        \end{align*}
        where the first inequality is due to Lemma \ref{lem:proj}.
    \end{proof}
    \begin{proof}[{\bf Proof of Lemma~\ref{lem:accuracy_glm}}]
        \label{proof_lem_accuracy_lowD}
        Note that by the proof of Lemma~\ref{lem:sensitivity_glm} we can see that
        when $n\geq \frac{16C_1^2t^2d}{\kappa_{2}^2}$,  with probability at least $1-2e^{-t^2d}$ we have $\|(\frac{X^TX}{n})^{-1}\|_{2}\leq \frac{2}{\kappa_{2}}$. Thus,
        \begin{align*}
            & \|\theta^{*}-\hat{\theta}(D)\|_{2}\\
            &=\|\theta^{*}-\big(\frac{X^TX}{n}\big)^{-1}\frac{X^T(A^{\prime})^{-1}(\proj(\widetilde{y}))}{n}\|_{2} \\
            & \leq \|\big(\frac{X^TX}{n}\big)^{-1}\|_{2} \|\big(\frac{X^TX}{n}\big)\theta^{*}-\frac{X^T(A^{\prime})^{-1}(\proj(\widetilde{y}))}{n}\|_{2} \\
            & \leq \frac{2}{\kappa_2}
            \|\frac{X^T}{n}\{X\theta^{*}-(A^{\prime})^{-1}(\proj(\widetilde{y}))\}\|_{2}\\
            &\leq \frac{2\sqrt{d}}{\kappa_{2}}\|\frac{X^T}{n}\{X\theta^{*}-(A^{\prime})^{-1}(\proj(\widetilde{y}))\}\|_{\infty}.
            \numberthis\label{lem:accuracy_lowD2}
        \end{align*}
        To complete the proof, we upper bound the term $\|\frac{X^T}{n}\{X\theta^{*}-(A^{\prime})^{-1}(\proj(\widetilde{y}))\}\|_{\infty}$.
        \begin{align*}
            & \quad  \|\frac{X^T}{n}\{X\theta^{*}-(A^{\prime})^{-1}(\proj(\widetilde{y}))\}\|_{\infty}\\
            & = \max_{j\in [d]} |[\frac{X^T}{n}\left\{X\theta^{*}-(A^{\prime})^{-1}(\proj(\widetilde{y}))\right\}]_j| \\
            & =\max_{j\in [d]} |\frac{1}{n}\sum_{i=1}^{n}x_{ij}(\sum_{k=1}^{d}x_{ik}\theta^{*}_{k}-(A^{\prime})^{-1}(\proj(\widetilde{y}_i)))| \\
            & =\max_{j\in [d]} |\frac{1}{n}\sum_{i=1}^{n}x_{ij}((A^{\prime})^{-1}\circ A^{\prime}(\langle \mathbf{x}_i,\theta^{*}\rangle)-
            (A^{\prime})^{-1}(\proj(\widetilde{y}_i)))| \\
            & = \max_{j\in [d]}|\frac{1}{n}\sum_{i=1}^{n}x_{ij}[(A^{\prime})^{-1}]^{\prime}(\xi_i)(A^{\prime}(\langle \mathbf{x}_i,\theta^{*}\rangle)-\proj(\widetilde{y}_i))| \quad \text{(by mean value theorem)}\\
            &\leq \max_{j\in [d]} |\frac{1}{n}\sum_{i=1}^{n}x_{ij}[(A^{\prime})^{-1}]^{\prime}(\xi_i)(A^{\prime}(\langle\mathbf{x}_i,\theta^{*}\rangle)-\widetilde{y}_i)|+\max_{j\in [d]}|\frac{1}{n}\sum_{i=1}^{n}x_{ij}[(A^{\prime})^{-1}]^{\prime}(\xi_i)(\widetilde{y}_i-\proj(\widetilde{y}_i))|\\
            &\equiv\max_{j\in [d]}\text{I}_j+\max_{j\in [d]}\text{II}_j,
            \numberthis\label{lem:accuracy_lowD3}
        \end{align*}
        where $\xi_i$ is some value between $A^{\prime}(\langle \mathbf{x}_i, \theta^{*}\rangle)$ and $\proj(\widetilde{y}_i)$.  Let $\mathcal{E}:=\{\max_{i\in [n]}\|\mathbf{x}_i\|_2\leq 4\sqrt{c}\sigma\sqrt{\log n}\}$ for constant $c>1$. By Lemma~\ref{lem:subG}, we have $\bb{P}(\mathcal{E}^c)\leq n^{-{c+1}}$. In the following, we will omit the conditioning on the event $\mathcal{E}$.
        
        Since $x_{ij}\stackrel{\mathrm{iid}}{\sim}\mathrm{subG}(\sigma^2/d), \forall i\in [n]$, by Lemma~\ref{lem:subG}, for any $t>0$, $\bb{P}(\text{I}_j>t)\leq 2\exp(-\frac{t^2}{2\sigma^2\|a\|_2^2/d})$, where $a=[\frac{1}{n}[(A^{\prime})^{-1}]^{\prime}(\xi_i)(A(\langle\mathbf{x}_i, \theta^{*}\rangle)-\widetilde{y}_i)]_{i=1}^n$. Let $t=c_1^{\prime}\sigma\|a\|_2\sqrt{\log n}/\sqrt{d}$. then
        \begin{align*}
            \bb{P}\left(\text{I}_j> c_1^{\prime}\frac{\sigma}{\sqrt{d}}\sqrt{\frac{1}{n}\sum_{i=1}^n([(A^{\prime})^{-1}]^{\prime}(\xi_i)(A^{\prime}(\langle\mathbf{x}_i,\theta^{*}\rangle)-\widetilde{y}_i))^2}\sqrt{\frac{\log n}{n}}\right)\leq 2n^{-\frac{c_1^{\prime^2}}{2}}.
        \end{align*}
        Since $|[(A^{\prime})^{-1}](\xi_i)|\leq \kappa_{A,0}$, with probability at least $1-2n^{-\frac{c_1^{\prime^2}}{2}}$ we have
        \begin{align*}
            \text{I}_j\leq  c_1^{\prime}\frac{\sigma}{\sqrt{d}}\kappa_{A,0}\sqrt{\frac{1}{n}\sum_{i=1}^n(A^{\prime}(\langle\mathbf{x}_i,\theta^{*}\rangle)-\widetilde{y}_i)^2}\sqrt{\frac{\log n}{n}}.
        \end{align*}
        Since $(A^{\prime}(\langle\mathbf{x}_i, \theta^{*}\rangle)-\widetilde{y}_i)^2\leq (M_A+\tau_2)^2$, by Hoeffding's inequality (Lemma~\ref{lem:Hoeffding}), with probability at least $1-2n^{-\frac{c_1^{\prime^2}}{2}}-2n^{-\zeta}$
        \begin{equation*}
            \text{I}_j\leq c_1'\frac{\sigma}{\sqrt{d}}\kappa_{A,0}\sqrt{\bb{E}[(A^{\prime}(\langle \mathbf{x}_{i},\theta^{*}\rangle)-\widetilde{y}_i)^2]+(M_A+\tau_2)^2\sqrt{\frac{\zeta\log n}{n}}}\sqrt{\frac{\log n}{n}}.
        \end{equation*}
        Note that
        \begin{align*}
            \bb{E}[(A^{\prime}(\langle \mathbf{x}_{i},\theta^{*}\rangle)-\widetilde{y}_i)^2]
            =\bb{E}[(A^{\prime}(\langle\mathbf{x}_i, \theta^{*}\rangle)-y_i+y_i-\widetilde{y}_i)^2]
            \leq 2(\bb{E}[(A^{\prime}(\langle\mathbf{x}_i, \theta^{*}\rangle)-y_i)^2]+
            \bb{E}[(y_i-\widetilde{y}_i)^2]).
            \numberthis\label{lem:accuracy_lowD4}
        \end{align*}
        Since $\bb{E}[y_i|\mathbf{x}_i,\theta^{*}]=A^{\prime}(\langle\mathbf{x}_i, \theta^{*}\rangle)$ and $\mathrm{var}[y_i|\mathbf{x}_i, \theta^{*}]=A^{\prime\prime}(\langle\mathbf{x}_i, \theta^{*}\rangle)\phi\leq \kappa_{A,2}\phi$, we have
        \begin{align*}
            \bb{E}[(A^{\prime}(\langle\mathbf{x}_i, \theta^{*}\rangle)-y_i)^2]
            =\bb{E}_{\mathbf{x}_i}[(\bb{E}_{y_i}[y_i|\mathbf{x}_i,\theta^{*}]-y_i)^2]
            =\bb{E}_{\mathbf{x}_i}[A^{\prime\prime}(\langle\mathbf{x}_i, \theta^{*}\rangle)\phi]
            \leq \kappa_{A,2}\phi.
            \numberthis\label{lem:accuracy_lowD5}
        \end{align*}
        For the second term of~\eqref{lem:accuracy_lowD4}, we have
        \begin{align*}
            \bb{E}[(y_i-\widetilde{y}_i)^2]
            =\bb{E}[(y_i-\widetilde{y}_i)^2\mathbf{1}(|y_i|>\tau_2)]
            &\leq \bb{E}[y_i^2\mathbf{1}_{\{|y_i|>\tau_2\}}]\\
            &\leq \sqrt{\bb{E}[y_i^4]\mathbb{P}(|y_i|>\tau_2)}
            \leq \sqrt{\bb{E}[y_i^4]}\sqrt{\frac{\bb{E}[y_i^4]}{\tau_2^4}}
            \leq \frac{R}{\tau_2^2}.
            \numberthis\label{lem:accuracy_lowD6}
        \end{align*}
        Combining~\eqref{lem:accuracy_lowD4},~\eqref{lem:accuracy_lowD5} and~\eqref{lem:accuracy_lowD6} delivers that
        \begin{align*}
            \text{I}_j\leq c_1^{\prime} \frac{\sigma}{\sqrt{d}}\kappa_{A,0}\sqrt{(\kappa_{A,2}\phi+\frac{R}{\tau_2^2})+(M_A+\tau_2)^2\sqrt{\frac{\zeta \log n}{n}}}\sqrt{\frac{\log n}{n}}. \numberthis\label{lem:accuracy_lowD7}
        \end{align*}
        Similarly, since $x_{ij}\stackrel{\mathrm{iid}}{\sim}\mathrm{subG}(\sigma^2/d)$, by Lemma \ref{lem:subG}, with probability at least $1-2n^{-\frac{c_2^{\prime^2}}{2}}$ it holds that
        \begin{align*}
            \text{II}_j\leq c_2^{\prime}\frac{\sigma}{\sqrt{d}}\kappa_{A,0}\varepsilon_{\bar{M}}\sqrt{\frac{\log n}{n}}.
            \numberthis\label{lem:accuracy_lowD8}
        \end{align*}
        Combining~\eqref{lem:accuracy_lowD7} and~\eqref{lem:accuracy_lowD8}, by the union bound for all $j\in [d]$, yields that with probability at least $1-2dn^{-\frac{c_1^{\prime^2}}{2}}-2dn^{-\zeta}-2dn^{-\frac{c_2^{\prime^2}}{2}}$,
        \begin{align*}
            \max_{j\in [d]}\text{I}_j+\max_{j\in [d]}\text{II}_j\leq O(( \sqrt{(\kappa_{A,2}\phi+\frac{R}{\tau_2^2})+(M_A+\tau_2)^2\sqrt{\frac{\zeta\log n}{n}}}+ \varepsilon_{\bar{M}})\frac{\sigma}{\sqrt{d
            }}\kappa_{A,0}\sqrt{\frac{\log n}{n}}).
            \numberthis\label{lem:accuracy_lowD9}
        \end{align*}
        Considering the failure of the event $\mathcal{E}$ and combining~\eqref{lem:accuracy_lowD2},~\eqref{lem:accuracy_lowD3}, and~\eqref{lem:accuracy_lowD9} delivers that for any $\delta>0$, with probability at least $1-n^{-c+1}-2dn^{-\frac{c_1^{\prime^2}}{2}}-2dn^{-\zeta}-2dn^{-\frac{c_2^{\prime^2}}{2}}=1-O(n^{-\Omega(1)})$,
        \begin{align*}
            \|\theta^{*}-(\frac{X^TX}{n})^{-1}\frac{X^T(A^{\prime})^{-1}(\proj(\widetilde{y}))}{n}\|_{2}\leq \lambda_n\sqrt{\frac{\log n}{n}}
        \end{align*}
        where $\lambda_n:=\tilde{O}({\kappa_{A,0}}(\sqrt{\kappa_{A,2}+\frac{1}{\tau_2^2}}+(M_A+\tau_2)\sqrt[4]{\frac{1}{n}}+\varepsilon_{\bar{\mathcal{M}}})$.
    \end{proof}
    \begin{proof}[{\bf Proof of Theorem~\ref{thm:privacy_glm}}]
        \label{proof_thm_privacy}
        We first show that $\hat{\theta}^P(\hat{D})$ is $(\varepsilon, \gamma_n)$-random joint differential privacy (RJDP). Let $\hat{D}$ and $\hat{D}^{\prime}$ be any two datasets that differ only on one agent's dataset. For any fixed $\theta\in \Theta\subseteq \bb{R}^d$, by Lemma~\ref{lem:subG_bound}, with probability at least $1-\gamma_n$ we have $\|\hat{\theta}(\hat{D})-\hat{\theta}(\hat{D}^{\prime})\|_2\leq \Delta_n$. Thus
        \begin{align*}
            &\quad \frac{p(\hat{\theta}^P(\hat{D})=\theta|\hat{D})}{p(\hat{\theta}^P(\hat{D}^{\prime})=\theta|\hat{D}^{\prime})}=\frac{p(\hat{\theta}(\hat{D})+v=\theta|\hat{D})}{p(\hat{\theta}(\hat{D}^{\prime})+v^{\prime}=\theta|\hat{D}^{\prime})} = \frac{p(v=\theta-\hat{\theta}(\hat{D})|\hat{D})}{p(v^{\prime}=\theta-\hat{\theta}(\hat{D}^{\prime})|\hat{D}^{\prime})} \\
            & =\exp\{\frac{\varepsilon}{\Delta_n}(\|\hat{\theta}(\hat{D})\|_2-\|\hat{\theta}(\hat{D}^{\prime})\|_2)\} \leq \exp\{\frac{\varepsilon}{\Delta_n}\|\hat{\theta}(\hat{D})-\hat{\theta}(\hat{D}^{\prime})\|_2\} \leq \exp(\varepsilon),
        \end{align*}
        which means $\hat{\theta}^P(\hat{D})$ is $(\varepsilon,\gamma_n)$-RJDP.
        The estimators $\hat{\theta}^P(\hat{D}^0)$ and $\hat{\theta}^P(\hat{D}^1)$ are computed in the same way as $\hat{\theta}^P(\hat{D})$, so $\hat{\theta}^P(\hat{D}^0)$ and $\hat{\theta}^P(\hat{D}^1)$ each satisfy $(\varepsilon,\gamma_{n/2})$-RJDP. Since $\hat{\theta}^P(\hat{D}^0)$ and $\hat{\theta}^P(\hat{D}^1)$ are computed on disjoint subsets of the data, then by the Parallel Composition Theorem, together they satisfy $(\varepsilon,2\gamma_{n/2})$-RJDP. By the Sequential Composition Theorem, the estimators ($\hat{\theta}^P(\hat{D})$,$\hat{\theta}^P(\hat{D}^0)$,$\hat{\theta}^P(\hat{D}^1)$) together satisfy $(2\varepsilon,\gamma_n+2\gamma_{n/2})$-RJDP.
        Finally, using  the post-processing property and Billboard Lemma~\ref{lem:billboard}, the output %$(\hat{\theta}^P(\hat{D})$,$\hat{\theta}^P(\hat{D}^0)$, $\hat{\theta}^P(\hat{D}^1)$, $\{\pi_i(D_i,\hat{\theta}^P(\hat{D}^b))\}_{i=1}^n)$
        $(\bar{\theta}^P(\hat{D})$,$\bar{\theta}^P(\hat{D}^0)$, $\bar{\theta}^P(\hat{D}^1)$, $\{\pi_i(D_i,\bar{\theta}^P(\hat{D}^b))\}_{i=1}^n)$
        of Mechanism \ref{mech:GLM} satisfies $(2\varepsilon,\gamma_n+2\gamma_{n/2})$-RJDP.
    \end{proof}
    
    \begin{proof}[{\bf Proof of Theorem~\ref{thm:truthfulness_glm}}]
        \label{proof_thm_truthfulness}
        Suppose all agents other than $i$ are following strategy $\sigma_{\tau_{\alpha, \beta}}$. Let agent $i$ be in group $1-b, b\in \{0,1\}$. We will show that $\sigma_{\tau_{\alpha, \beta}}$ achieves $\eta$-Bayesian Nash equilibrium by bounding agent $i$'s incentive to deviate. Assume that $c_i\leq \tau_{\alpha,\beta}$, otherwise there is nothing to show because agent $i$ would be allowed to submit an arbitrary report under $\sigma_{\tau_{\alpha, \beta}}$. For ease of notation,  we write $\sigma$ for $\sigma_{\tau_{\alpha,\beta}}$ for the remainder of the proof. We first compute the maximum expected  mount (based on his belief) that agent $i$ can increase his payment by misreporting to the analyst, i.e.
        \begin{align*}
            & \quad  \bb{E}[\pi_i(\hat{D}_i, \sigma(D^b, c^b))|D_i,c_i]-\bb{E}[\pi_i(D_i, \sigma(D^b, c^b))|D_i,c_i]\\
            & =\bb{E}\left[B_{a_1,a_2}\left(A^{\prime}(\langle  \mathbf{x}_i,\bar{\theta}^{P}(\hat{D}^{b})\rangle),A^{\prime}(\langle \mathbf{x}_i, \bb{E}_{\theta\sim p(\theta|\hat{D}_i)}[\theta] \rangle)\right)\big| D_i,c_i\right]\\
            &\quad  - \bb{E}\left[B_{a_1,a_2}\left(A^{\prime}(\langle  \mathbf{x}_i,\bar{\theta}^{P}(\hat{D}^{b})\rangle),A^{\prime}(\langle \mathbf{x}_i, \bb{E}_{\theta\sim p(\theta|D_i)}[\theta] \rangle)\right)\big| D_i,c_i\right]. \numberthis \label{thm:truthfulness1}
        \end{align*}
        Note that $B_{a_1,a_2}(p,q)=a_1-a_2(p-2pq+q^2)$ is linear with respect to $p$, and is a strictly concave function of $q$ maximized at $q=p$. Thus,~\eqref{thm:truthfulness1} is upper bounded by the following with probability $1-C_1n^{-\Omega(1)}$
        \begin{align*}
            &\quad B_{a_1,a_2}(\bb{E}[A^{\prime}(\langle \mathbf{x}_i,\bar{\theta}^P(\hat{D}^b)\rangle)|D_i, c_i], \bb{E}[A^{\prime}(\langle \mathbf{x}_i,\hat{\theta}^P(\hat{D}^b)\rangle)|D_i, c_i])\\
            &\quad -
            B_{a_1,a_2}(\bb{E}[A^{\prime}(\langle  \mathbf{x}_i,\bar{\theta}^{P}(\hat{D}^{b})\rangle)|D_i,c_i],A^{\prime}(\langle \mathbf{x}_i, \bb{E}_{\theta\sim p(\theta|D_i)}[\theta] \rangle))\\
            & = a_2\left(\bb{E}[A^{\prime}(\langle \mathbf{x}_i,\bar{\theta}^P(\hat{D}^b)\rangle)|D_i,c_i]-A^{\prime}(\langle \mathbf{x}_i, \bb{E}_{\theta\sim p(\theta|D_i)}[\theta] \rangle)\right)^2\\
            & =a_2\left(\bb{E}[A^{\prime}(\langle \mathbf{x}_i,\bar{\theta}^P(\hat{D}^b)\rangle)-A^{\prime}(\langle \mathbf{x}_i, \bb{E}_{\theta\sim p(\theta|D_i)}[\theta] \rangle)|D_i,c_i]\right)^2\\
            & \leq a_2\left(\bb{E}[\kappa_{A,2}\mathbf{x}_i^T(\bar{\theta}^P(\hat{D}^b)-\bb{E}_{\theta\sim p(\theta|D_i)}[\theta])|D_i,c_i]\right)^2\\
            & \leq a_2\kappa_{A,2}^2\|\mathbf{x}_i\|_2^2\|\bb{E}[\bar{\theta}^P(\hat{D}^b)-\bb{E}_{\theta\sim p(\theta|D_i)}[\theta]|D_i,c_i]\|_2^2 \\
            & \leq C a_2\kappa_{A,2}^2\sigma^2\log n\|\bb{E}[\bar{\theta}^P(\hat{D}^b)-\bb{E}_{\theta\sim p(\theta|D_i)}[\theta]|D_i,c_i]\|_2^2.
        \end{align*}
        We continue by bounding the term $\|\bb{E}[\bar{\theta}^P(\hat{D}^b)-\bb{E}_{\theta\sim p(\theta|D_i)}[\theta]|D_i,c_i]\|_2$. By Lemma \ref{lem:proj}
        \begin{align*}
            &\|\bb{E}[\bar{\theta}^P(\hat{D}^b)-\bb{E}_{\theta\sim p(\theta|D_i)}[\theta]|D_i,c_i]\|_2 \\
            &\leq \|\bb{E}[\bar{\theta}^P(\hat{D}^b)-\bar{\theta}^P(D^b)|D_i, c_i]\|_2+ \| \bb{E}[\bar{\theta}^P(D^b)|D_i, c_i]-\bb{E}_{\theta\sim p(\theta|D_i)}[\theta]|D_i,c_i]\|_2 \\
            &\leq \|\bb{E}[\hat{\theta}^P(\hat{D}^b)-\hat{\theta}^P(D^b)|D_i, c_i]\|_2+ \| \bb{E}[\bar{\theta}^P(D^b)|D_i, c_i]-\bb{E}_{\theta\sim p(\theta|D_i)}[\theta]|D_i,c_i]\|_2 \\
            %&=\|\bb{E}[\hat{\theta}(\hat{D}^b)+v_b|D_i,c_i]-\bb{E}_{\theta\sim p(\theta|D_i)}[\theta]\|_2\\
            %&=\|\bb{E}[\hat{\theta}(\hat{D}^b)|D_i, c_i]-\bb{E}_{\theta\sim p(\theta|D_i)}[\theta]\|_2 \\
            %&=\|\bb{E}[\hat{\theta}(\hat{D}^b)|D_i]-\bb{E}_{\theta\sim p(\theta|D_i)}[\theta]\|_2 \\
            %&=\|\bb{E}[\hat{\theta}(\hat{D}^b)-\hat{\theta}(D^b)|D_i]+\bb{E}[\hat{\theta}(D^b)|D_i]-\bb{E}_{\theta\sim p(\theta|D_i)}[\theta]\|_2\\
            & \leq \bb{E}\|\hat{\theta}(\hat{D}^b)-\hat{\theta}(D^b)\|_2+
            \|\bb{E}[\bar{\theta}^P(D^b)|D_i]-\bb{E}_{\theta\sim p(\theta|D_i)}[\theta]\|_2,
            \numberthis \label{thm:truthfulness2}
        \end{align*}
        %where the second equality holds since $v_b$ is independent of $D_i, c_i$ and $\bb{E}[v_b]=0$;  the  third equality uses Assumption~\ref{assump:privacy_cost_independence}.
        Since agent $i$ believes that with at least probability $1-\beta$, at most $\alpha n$ agents will misreport their datasets under threshold strategy $\sigma_{\tau_{\alpha, \beta}}$, datasets $D^b$ and $\hat{D}^b$ differ only on at most $\alpha n$ agents' datasets.
        By Lemma~\ref{lem:sensitivity_k_entries}, with probability at least $1-\alpha n \gamma_{n/2}$ we have $\bb{E}\|\hat{\theta}(\hat{D}^b)-\hat{\theta}(D^b)\|_2\leq \alpha n\Delta_{n/2}$. For the third term of ~\eqref{thm:truthfulness2},
        \begin{align*}
            &\quad\bb{E}[\bar{\theta}^P(D^b)|D_i]-\bb{E}_{\theta\sim p(\theta|D_i)}[\theta]\\
            & =\bb{E}_{D^b\sim p(D^b|D_i)}[\bar{\theta}^P(D^b)]-\bb{E}_{\theta\sim p(\theta|D_i)}[\theta]\\
            & =\bb{E}_{\theta \sim p(\theta|D_i)}[\bb{E}_{D^b\sim p(D^b|\theta)}[\bar{\theta}^P(D^b)]|\theta]-\bb{E}_{\theta\sim p(\theta|D_i)}[\theta]\\
            & = \bb{E}_{\theta\sim p(\theta|D_i)}[\bb{E}_{D^b\sim p(D^b|\theta)}[\bar{\theta}^P(D^b)-\theta]|\theta].
        \end{align*}
        Since
        \begin{align*}
            p(D^b|\theta)=p(X^b,y^b|\theta)=p(y^b|X^b,\theta)p(X^b|\theta)=p(y^b|X^b,\theta)p(X^b),
        \end{align*}
        we have
        \begin{align*}
            \bb{E}_{D^b\sim p(D^b|\theta)}[\hat{\theta}(D^b)-\theta]=\bb{E}_{X^b}[\bb{E}_{y^b}[\bar{\theta}^P(X^b,y^b)-\theta]| X^b, \theta].
        \end{align*}
        Since we have the prior knowledge that $\|\theta^*\|_2\leq \tau_\theta$. Thus, the for posterior distribution  $\theta\sim p(\theta|\hat{D}_i)$  it will also have $\|\theta\|_2\leq \tau_\theta$.
        By Jensen's inequality, Lemma~\ref{lem:proj} and Lemma~\ref{lem:accuracy_glm}, with probability at least $1-O(n^{-\Omega(1)})$ we have
        \begin{align*}
            &\|\bb{E}[\bar{\theta}^P(D^b)|D_i]-\bb{E}_{\theta\sim p(\theta|D_i)}[\theta]\|_2
            \leq \bb{E}_{\theta\sim p(\theta |D_i), X^b}[\bb{E}_{y^b}[\|\bar{\theta}^P(X^b,y^b)-\theta\|_2|X^b, \theta]]\\
            &  \leq \bb{E}_{\theta\sim p(\theta |D_i), X^b}[\bb{E}_{y^b}[\|\hat{\theta}^P(X^b,y^b)-\theta\|_2|X^b, \theta]]
            \leq \lambda_n\sqrt{\frac{\log n}{n}}+\mathbb{E}\|v_b\|_2,
        \end{align*}
        where $ \lambda_n:=\tilde{O}({\sigma\kappa_{A,0}}(\sqrt{\kappa_{A,2}+\frac{1}{\tau_2^2}}+(M_A+\tau_2)\sqrt[4]{\frac{1}{n}}+\varepsilon_{\bar{\mathcal{M}}}).$
        In addition to an increased payment, agent $i$ may also experience decreased privacy costs from misreporting. By Assumption~\ref{assump2:privacy_cost_function_bound}, this decrease in privacy costs is bounded above by $c_iF(2\varepsilon, \gamma_n+2\gamma_{n/2})$. Since we have assumed $c_i\leq \tau_{\alpha,\beta}$, the decrease in privacy costs for agent $i$ is bounded above by $\tau_{\alpha,\beta}F(2\varepsilon, \gamma_n+2\gamma_{n/2})$. Hence, agent $i$'s total incentive to deviate is bounded above by
        \begin{align*}
            \eta =O(a_2\kappa_{A,2}^2\sigma^2\log n(\alpha n\Delta_{n/2} +\lambda_n\sqrt{\frac{\log n}{n}}+\frac{d\Delta_{n/2}}{\epsilon})^2+\tau_{\alpha,\beta}F(2\varepsilon, \gamma_n+2\gamma_{n/2})).
        \end{align*}
    \end{proof}
    \begin{proof}[{\bf Proof of Theorem~\ref{thm:accuracy_glm}}]
        For any realization $D$ held by agents, let $\hat{D}=\sigma_{\tau_{\alpha, \beta}}(D)$.  Then by Lemma \ref{lem:proj} we have
        \begin{align*}
            &\bb{E}[\|\bar{\theta}^P(\hat{D})-\theta^{*}\|_2^2]\leq   \bb{E}\|\hat{\theta}^P(\hat{D})-\theta^{*}\|_2^2\\
            & = \bb{E}\|\hat{\theta}^P(\hat{D})-\hat{\theta}(D)+\hat{\theta}(D)-\theta^{*}\|_2^2\\
            & = \bb{E}\|\hat{\theta}^P(\hat{D})-\hat{\theta}(D)\|_2^2+2\langle \hat{\theta}^P(\hat{D})-\hat{\theta}(D), \hat{\theta}(D)-\theta^{*}\rangle++\|\hat{\theta}(D)-\theta^{*}\|_2^2 \\
            & \leq  2\bb{E}\|\hat{\theta}^P(\hat{D})-\hat{\theta}(D)\|_2^2 +
            2\bb{E}\|\hat{\theta}(D)-\theta^{*}\|_2^2. \numberthis\label{thm:accuracy1}
        \end{align*}
        For the first term of~\eqref{thm:accuracy1}, by Lemma~\ref{lem:sensitivity_k_entries} and Lemma~\ref{lem:moment}, with probability at least $1-\beta-\alpha n\gamma_n$,  we have
        \begin{align*}
            \bb{E}\|\hat{\theta}^P(\hat{D})-\hat{\theta}(D)\|_2^2 &=\bb{E}\|\hat{\theta}(\hat{D})+v-\hat{\theta}(D)\|_2^2\\
            &=  \bb{E}\|\hat{\theta}(\hat{D})-\hat{\theta}(D)\|_2^2 + \bb{E}\|v\|_2^2+2\bb{E}\langle\hat{\theta}(\hat{D})-\hat{\theta}(D),v\rangle\\
            & = \bb{E}\|\hat{\theta}(\hat{D})-\hat{\theta}(D)\|_2^2 + \bb{E}\|v\|_2^2+2\langle\bb{E}\hat{\theta}(\hat{D})-\hat{\theta}(D),\bb{E}[v]\rangle \\
            & \leq (\alpha n \Delta_n)^2+ d(d+1)(\frac{\Delta_n}{\varepsilon})^2.
            \numberthis\label{thm:accuracy2}
        \end{align*}
        For the last term of~\eqref{thm:accuracy1}, by Lemma~\ref{lem:accuracy_glm},
        \begin{align*}
            \|\hat{\theta}(D)-\theta^{*}\|^2_2\leq \lambda^2_n {\frac{\log n}{n}}. \numberthis\label{thm:accuracy4}
        \end{align*}
        Combining~\eqref{thm:accuracy2} and~\eqref{thm:accuracy4} yields that with probability at least $1-\beta-Cdn^{-\Omega(1)}$,
        \begin{align*}
            \bb{E}[\|\hat{\theta}^P(\hat{D})-\theta^{*}\|_2^2]\leq
            O( (\alpha n \Delta_n)^2+ d^2(\frac{\Delta_n}{\varepsilon})^2 +\lambda^2_n {\frac{\log n}{n}}).
        \end{align*}
    \end{proof}

    \begin{proof}[\bf Proof of Theorem~\ref{thm:rationality_glm}]
        Let agent $i$ have privacy cost $c_i\leq \tau_{\alpha,\beta}$ and consider agent $i$'s utility from participating in the mechanism. Suppose agent $i$ is in group $1-b$, then his expected utility is
        \begin{align*}
            \bb{E}[u_i]
            &=\bb{E}\left[B_{a_1,a_2}\left(A^{\prime}(\langle  \mathbf{x}_i,\bar{\theta}^{P}(\hat{D}^{b})\rangle),A^{\prime}(\langle \mathbf{x}_i, \bb{E}_{\theta\sim p(\theta|\hat{D}_i)}[\theta] \rangle)\right) |D_i, c_i\right]-f_{i}(c_i,\varepsilon)\\
            & \geq B_{a_1,a_2}\left(\bb{E}[A^{\prime}(\langle  \mathbf{x}_i,\bar{\theta}^{P}(\hat{D}^{b})\rangle)|D_i,c_i],A^{\prime}(\langle \mathbf{x}_i, \bb{E}_{\theta\sim p(\theta|\hat{D}_i)}[\theta] \rangle)\right)-\tau_{\alpha, \beta}F(2\varepsilon, \gamma_n+2\gamma_{n/2}).
            \numberthis\label{thm:rationality1}
        \end{align*}
        Since both $|\langle\mathbf{x}_i, \bar{\theta}^P(\hat{D}^b)\rangle|$ and $|\langle \mathbf{x}_i, \bb{E}_{\theta\sim p(\theta|\hat{D}_i)}[\theta]\rangle|$ are bounded by $\tau_1\tau_{\theta}$ with probability at least $1-\beta-O(n^{-\Omega(1)})$,  both two inputs of $B_{a_1, a_2}(\cdot, \cdot)$ are bounded above by $M_A$. Note that
        \begin{align*}
            B_{a_1,a_2}(p,q)=a_1-a_2(p-2pq+q^2)\geq a_1-a_2(|p|+2|p||q|+|q|^2),
            \numberthis\label{thm:rationality2}
        \end{align*}
        thus by~\eqref{thm:rationality1} and~\eqref{thm:rationality2} agent $i$'s expected utility is non-negative as long as
        \begin{align*}
            a_1\geq a_2(M_A+3M_A^2)+\tau_{\alpha,\beta}F(2\varepsilon, \gamma_n+\gamma_{n/2}).
        \end{align*}
    \end{proof}

    \begin{proof}[{\bf Proof of Theorem~\ref{thm:budget_glm}}]
        Note that
        \begin{align*}
            B_{a_1,a_2}(p,q)\leq B_{a_1,a_2}(p,p)=a_1-a_2(p-p^2)\leq a_1+a_2(|p|+|p|^2),
        \end{align*}
        thus
        \begin{align*}
            \mathcal{B}=\sum_{i=1}^{n}\bb{E}[\pi_i]
            &=\sum_{i=1}^{n}\bb{E}[B_{a_1,a_2}\left(A^{\prime}(\langle\mathbf{x}_i,\bar{\theta}^{P}(\hat{D}^{b})\rangle),A^{\prime}(\langle \mathbf{x}_i, \bb{E}_{\theta\sim p(\theta|\hat{D}_i)}[\theta] \rangle)\right)|D_i,c_i]\\
            &\leq n(a_1+a_2(M_A+M_A^2)).
        \end{align*}
    \end{proof}
    
    \begin{proof}[{\bf Proof of Corollary~\ref{cor:glm_linear}}]
        The response moment polytope for the real-valued response variable is $\mathcal{M}=\bb{R}$. Thus its interior is $\mathcal{M}^{\circ}=\bb{R}$. We set $\bar{\mathcal{M}}=\bb{R}$, so that $\proj(\widetilde{y}_i)=\widetilde{y}_i, \varepsilon_{\bar{\mathcal{M}}}=0$. To bound $\kappa_{A,0}, \kappa_{A,1}, \kappa_{A,2}, M_A$, we first compute that $A^{\prime}(a)=a$, $(A^{\prime})^{-1}(a)=a$, $[(A^{\prime})^{-1}]^{\prime}(a)=1$, $A^{\prime\prime}(a)=1$. Note that $\mathcal{M}^{\prime}=[-\tau_{\theta}\tau_1, \tau_{\theta}\tau_1]=[-C\sigma\tau_{\theta}\sqrt{\log n}, C\sigma\tau_{\theta}\sqrt{\log n}]$, thus we have $$\kappa_{A,0}=1, \kappa_{A,1}=\tau_2, \kappa_{A,2}=1, M_A=C\sigma\tau_{\theta}\sqrt{\log n}.$$ 
        
        For any $\delta\in (\frac{1}{4}, \frac{1}{3})$ and $c>0$, we set $\varepsilon=n^{-\delta}$, $\tau_2=n^{\frac{1-3\delta}{2}}$, $\alpha=\Theta(n^{-3\delta})$, $\beta=\Theta(n^{-c})$ , $a_2=O(n^{-4\delta})$, and $a_1=a_2(M_A+3M_A^2)+\tau_{\alpha,\beta}F(2\varepsilon, \gamma_n+2\gamma_{n/2})$. Then, by Lemma~\ref{lem:sensitivity_glm}, the sensitivity of the private estimator is $\Delta_n=\widetilde{O}(\tau_2n^{-\frac{1}{2}})=\widetilde{O}(n^{-\frac{3\delta}{2}})$. Since for any $\delta\in (\frac{1}{4}, \frac{1}{3})$ we always have
        $\frac{1-3\delta}{2}<\frac{1}{4}$, which means $\tau_2=O(n^{\frac{1}{4}})$, by Lemma~\ref{lem:accuracy_glm} we obtain  $\lambda_n=O(1)$, $\|\bar{\theta}(D)-\theta^{*}\|_2=\widetilde{O}(n^{-\frac{1}{2}})$.
         
         Recall that by Theorem~\ref{thm:accuracy_glm}, the private estimator is $\widetilde{O}(\alpha^2\kappa_{A,1}^2nd+\frac{\kappa_{A,1}^2d^3}{\varepsilon^2n}+\frac{\lambda_n^2}{n})$-accurate. Note that 
         $\widetilde{O}(\alpha^2\kappa_{A,1}^2nd)=\widetilde{O}(n^{-9\delta+2})$, $\widetilde{O}(\frac{\kappa_{A,1}^2d^3}{\varepsilon^2 n})=\widetilde{O}(n^{-\delta})$, $\widetilde{O}(\frac{\lambda_n^2}{n})=\widetilde{O}(n^{-1})$. Since for any $\delta\in (\frac{1}{4}, \frac{1}{3})$ it holds that $-1<-9\delta+2<-\delta$, we obtain $\bb{E}\|\bar{\theta}^P(\hat{D})-\theta^{*}\|_2^2=\widetilde{O}(n^{-\delta})$. To bound the expected budget, we first bound the threshold value $\tau_{\alpha, \beta}$ and the term $\tau_{\alpha, \beta}F(2\varepsilon, \gamma_n+2\gamma_{n/2})$. By Lemma~\ref{lem:bound_on_threshold}, $\tau_{\alpha,\beta}\leq \frac{1}{\lambda}\log \frac{1}{\alpha\beta}=\Theta(\frac{3\delta+c}{\lambda}\log n)=
         \widetilde{\Theta}(1)$. If $F(\varepsilon, \gamma)=(1+\gamma)\varepsilon^4$, then $\tau_{\alpha,\beta}F(2\varepsilon, \gamma_n+2\gamma_{n/2})=\widetilde{O}(n^{-4\delta})$. Recall that by Theorem~\ref{thm:truthfulness_glm}, the first term of truthfulness bound is $a_2\kappa_{2}^2(\alpha^2\kappa_{A, 1}^2nd+\frac{\lambda_n^2}{n}+\frac{\kappa_{A,1}^2d^3}{n\varepsilon^2})=\widetilde{O}(n^{-5\delta})$, thus $\eta=\widetilde{O}(n^{-5\delta}+n^{-4\delta})=\widetilde{O}(n^{-4\delta})$. By the choice of $a_1$ and Theorem~\ref{thm:rationality_glm}, the mechanism is individual rational for at least $1-O(n^{-3\delta})$ fraction of agents.
        By Theorem~\ref{thm:budget_glm}, the total expected budget is $\mathcal{B}=\widetilde{O}(na_2(2M_A+4M_A^2)+n\tau_{\alpha,\beta}F(2\varepsilon, \gamma_n+2\gamma_{n/2}))=
        \widetilde{O}(n^{-4\delta+1})$.
    \end{proof}
    
    \begin{proof}[{\bf Proof of Corollary~\ref{cor:glm_logistic}}]
         The response moment polytope for the binary response variable $y\in \{-1, 1\}$ is
        $\mathcal{M}=[-1,1]$. Thus,  its interior is given by $\mathcal{M}^{\circ}=(-1,1)$. For the closed subset of $\mathcal{M}^{\circ}$, we define it as $\bar{\mathcal{M}}=[-1+\varepsilon^{\prime}, 1-\varepsilon^{\prime}]$ for some $\varepsilon^{\prime}\in (0,1)$. Then, we can easily compute that $\proj(\widetilde{y}_i)=\widetilde{y}_i(1-\varepsilon^{\prime})$, $\varepsilon_{\bar{\mathcal{M}}}=\varepsilon^{\prime}$.  Since $|y|\leq 1$, here we can just set $\tau_2=1$.
        To bound $\kappa_{A,0}, \kappa_{A,1}, \kappa_{A,2}, M_A$, we first compute that $A^{\prime}(a)=\frac{e^{2a}-1}{e^{2a}+1}$, $(A^{\prime})^{-1}(a)=\frac{1}{2}\log \frac{1+a}{1-a}$. $[(A^{\prime})^{-1}]^{\prime}(a)=\frac{1}{1-a^2}$, $A^{\prime\prime}(a)=\frac{4}{(e^{a}+e^{-a})^2}$. Note that $\mathcal{M}^{\prime}=[-\frac{e^{2C\sigma\tau_{\theta}\sqrt{\log n}}-1}{e^{2C\sigma\tau_{\theta}\sqrt{\log n}}+1}, \frac{e^{2C\sigma\tau_{\theta}\sqrt{\log n}}-1}{e^{2C\sigma\tau_{\theta}\sqrt{\log n}}+1}]$, then we have 
        \begin{align*}
            &\kappa_{A,0}=\max_{a\in \mathcal{M}^{\prime}\cup \bar{\mathcal{M}}}|[(A^{\prime})^{-1}]^{\prime}(a)|=\max_{a\in \mathcal{M}^{\prime}\cup \bar{\mathcal{M}}}\frac{1}{2}(\frac{1}{1-a}+\frac{1}{1+a})\\
            &< \max\{\frac{1}{2}+\frac{1}{2}e^{2C\sigma\tau_{\theta}\sqrt{\log n}}, \frac{1}{\varepsilon^{\prime}}\}=\max\{\frac{1}{2}+\frac{1}{2}n^{\frac{2C\sigma\tau_{\theta}}{\sqrt{\log n}}},\frac{1}{\varepsilon^{\prime}}\},\\
            &\kappa_{A,1}=\frac{1}{2}\log \frac{2-\varepsilon^{\prime}}{\varepsilon^{\prime}}, \kappa_{A,2}\leq 1, M_A\leq 1.
        \end{align*}
    
        For any $\delta\in (\frac{1}{4}, \frac{1}{2})$, we choose $\varepsilon^{\prime}=2n^{-\delta}$ and let $n\geq e^{(\frac{2C\sigma\tau_{\theta}}{\delta})^2}$ then $\kappa_{A,0}=O(n^{\delta})$, $\kappa_{A,1}=\widetilde{O}(1)$. We
        set $\varepsilon=n^{-\delta}$, $\alpha=\Theta(n^{-3\delta})$, $\beta=\Theta(n^{-c})$ for any $c>0$, $a_2=n^{-4\delta}$, $a_1=a_2(M_A+3M_A^2)+\tau_{\alpha,\beta}F(2\varepsilon, \gamma_n+2\gamma_{n/2})$. Then, by Lemma~\ref{lem:sensitivity_glm}, the sensitivity of the private estimator is  $\Delta_n=\widetilde{O}(n^{-\frac{1}{2}})$. Then, by Lemma~\ref{lem:accuracy_glm}, we have  $\lambda_n=\widetilde{O}(\kappa_{A,0})=\widetilde{O}(n^{\delta})$ and $\|\bar{\theta}(D)-\theta^{*}\|_2 =\widetilde{O}(n^{-\frac{1-2\delta}{2}})$. Note that
        $\alpha^2\kappa_{A, 1}^2nd=\widetilde{O}(n^{-6\delta+1})$, $\frac{\kappa_{A, 1}^2d^3}{\varepsilon^2n}=n^{-1+2\delta}$, $\frac{\lambda_n^2}{n}=\widetilde{O}(n^{-1+2\delta})$. Since for any $\delta\in (\frac{1}{4}, \frac{1}{2})$ it holds that $-6\delta+1<-1+2\delta$, by Theorem~\ref{thm:accuracy_glm}, we obtain $\bb{E}[\|\bar{\theta}^P(\hat{D})-\theta^{*}\|_2^2]=\widetilde{O}(n^{-1+2\delta})$. By Lemma~\ref{lem:bound_on_threshold} and the assumption that $F(\varepsilon, \gamma)= (1+\gamma)\varepsilon^4$, $\tau_{\alpha,\beta}F(2\varepsilon, \gamma_n+2\gamma_{n/2})=\widetilde{O}(n^{-4\delta})$. 
        Note that $a_2\kappa_{A,2}^2(\alpha^2\kappa_{A, 1}^2nd+\frac{\lambda_n^2}{n}+\frac{\kappa_{A, 1}^2d^3}{n\varepsilon^2})=\widetilde{O}(n^{-1-2\delta})=\widetilde{O}(n^{-4\delta})$, thus by Theorem~\ref{thm:truthfulness_glm}, $\eta=\widetilde{O}(n^{-4\delta})$. By the choice of $a_1$ and Theorem~\ref{thm:rationality_glm}, the mechanism is individual rational for at least $1-O(n^{-3\delta})$ fraction of agents.
        By Theorem~\ref{thm:budget_glm}, the total expected budget is $\mathcal{B}=\widetilde{O}(na_2(2M_A+4M_A^2)+n\tau_{\alpha,\beta}F(2\varepsilon, \gamma_n+2\gamma_{n/2}))=
        \widetilde{O}(n^{-4\delta+1})$.
    \end{proof}
    
    \begin{proof}[{\bf Proof of Corollary~\ref{cor:glm_poisson}}]
        In this case, the response moment polytope for the count-valued $y\in \{0,1,2,\cdots\}$ is $\mathcal{M}=[0,+\infty)$. Thus, its interior is given by $\mathcal{M}^{\circ}=(0,+\infty)$. For the closed subset of the interior, we define $\bar{\mathcal{M}}=[\varepsilon^{\prime}, +\infty)$, for some $\varepsilon^{\prime}\in (0,1)$, and thus $\proj(\widetilde{y}_i)=\mathbf{1}_{\{\widetilde{y}_i=0\}}\varepsilon^{\prime}+\mathbf{1}_{\{\widetilde{y}_i\neq 0\}}\widetilde{y}_i$, $\varepsilon_{\bar{\mathcal{M}}}=\varepsilon^{\prime}$. To bound $\kappa_{A,0}, \kappa_{A,1}, \kappa_{A,2}, M_A$, we compute that $A^{\prime}(a)=e^a$, $(A^{\prime})^{-1}(a)=\log a$, $[(A^{\prime})^{-1}]^{\prime}(a)=\frac{1}{a}$, $A^{\prime\prime}(a)=e^a$. Note that $\mathcal{M}^{\prime}=[e^{-C\sigma\tau_{\theta}\sqrt{\log n}}, e^{C\sigma\tau_{\theta}\sqrt{\log n}}]$, then we have 
        \begin{align*}
            &\kappa_{A,0}=\max_{a\in \mathcal{M}^{\prime}\cup\bar{\mathcal{M}}}|[(A^{\prime})^{-1}]^{\prime}(a)|=\max_{a\in \mathcal{M}^{\prime}\cup \bar{\mathcal{M}}}|\frac{1}{a}|=\max\{e^{C\sigma\tau_{\theta}\sqrt{\log n}}, \frac{1}{\varepsilon^{\prime}}\}=\max\{n^{\frac{C\sigma\tau_{\theta}}{\sqrt{\log n}}}, \frac{1}{\varepsilon^{\prime}}\} \\
            & \kappa_{A,1}=\max\{|\log \varepsilon^{\prime}|, |\log \tau_2|\},\quad  \kappa_{A,2}=M_A=n^{\frac{C\sigma\tau_{\theta}}{\sqrt{\log n}}}.
        \end{align*}
    
        For any $\delta\in (\frac{1}{4}, \frac{1}{3})$, set $\varepsilon^{\prime}=n^{-\delta}$, then when 
        $n\geq e^{(\frac{C\sigma\tau_{\theta}}{\delta})^2}$, we have $\kappa_{A,0}=O(n^{\delta})$. 
        Also  $\kappa_{A,2}=M_A=O(n^{\delta})$. We choose $\varepsilon=n^{-3\delta}$, $\tau_2=\Theta(n^{\frac{1}{4}})$, $\alpha=\Theta(n^{-3\delta})$, , $\beta=\Theta(n^{-c})$ for any $c>0$, $a_2=n^{-6\delta}$, and $a_1=a_2(M_A+3M_A^2)+\tau_{\alpha,\beta}F(2\varepsilon, \gamma_n+2\gamma_{n/2})$. By Lemma~\ref{lem:sensitivity_glm}, the sensitivity of the private estimator is  $\Delta_n=\widetilde{O}(n^{-\frac{1}{2}})$. 
        Then, recall that by Lemma~\ref{lem:accuracy_glm},
        $\lambda_n=\widetilde{O}(\kappa_{A,0}(\sqrt{\kappa_{A,2}+\frac{1}{\tau_2^2}}+(M_A+\tau_2)\sqrt[4]{\frac{1}{n}}+\varepsilon_{\bar{\mathcal{M}}}))$. We have $\sqrt{\kappa_{2}+\frac{1}{\tau_2^2}}=O(n^{\frac{\delta}{2}})$, 
        $(M_A+\tau_2)\sqrt[4]{\frac{1}{n}}=O(n^{-\frac{1}{4}+\delta})=O(n^{\frac{\delta}{2}})$, $\varepsilon_{\bar{\mathcal{M}}}=O(1)$. Thus,  $\lambda_n=\widetilde{O}(n^{\frac{3\delta}{2}})$ and $\|\bar{\theta}(D)-\theta^{*}\|_2=\widetilde{O}(n^{-\frac{1-3\delta}{2}})$. Nota that $\alpha^2\kappa_{A, 1}^2nd=\widetilde{O}(n^{-6\delta+1})$, $\frac{\kappa_{A, 1}^2d^3}{\varepsilon^2n}=\widetilde{O}(n^{-1+2\delta})$, $\frac{\lambda_n^2}{n}=\widetilde{O}(n^{-1+3\delta})$. For any $\delta\in (\frac{1}{4}, \frac{1}{3})$, it holds that $-6\delta+1<-1+2\delta<-1+3\delta$, thus by Theorem~\ref{thm:accuracy_glm}, we obtain $\bb{E}\|\hat{\theta}^P(\hat{D})-\theta^{*}\|_2^2=\widetilde{O}(n^{-1+3\delta})$. By Lemma~\ref{lem:bound_on_threshold} and the assumption that $F(\varepsilon, \gamma)= (1+\gamma)\varepsilon^4$, $\tau_{\alpha,\beta}F(2\varepsilon, \gamma_n+2\gamma_{n/2})=\widetilde{O}(n^{-4\delta})$. 
        Note that $a_2\kappa_{A,2}^2(\alpha^2\kappa_{A, 1}^2nd+\frac{\lambda_n^2}{n}+\frac{\kappa_{A, 1}^2d^3}{n\varepsilon^2})=\widetilde{O}(n^{-1-\delta})=\widetilde{O}(n^{-4\delta})$, thus by Theorem~\ref{thm:truthfulness_glm}, $\eta=\widetilde{O}(n^{-4\delta})$. By the choice of $a_1$ and Theorem~\ref{thm:rationality_glm}, the mechanism is individual rational for at least $1-O(n^{-3\delta})$ fraction of agents.
        By Theorem~\ref{thm:budget_glm}, the total expected budget is $\mathcal{B}=\widetilde{O}(na_2(2M_A+4M_A^2)+n\tau_{\alpha,\beta}F(2\varepsilon, \gamma_n+2\gamma_{n/2}))=
        \widetilde{O}(n^{-4\delta+1})$.
    \end{proof}
    
    \begin{proof}[{\bf Proof of Lemma~\ref{lem:sensitivity_linear}}]
        \label{proof_lem_sensitivity_linear}
        We apply the same techniques as in the proof of Lemma~\ref{lem:sensitivity_glm}.
        Let $D$ and $D^{\prime}$ be two arbitrary neighboring datasets that differ only on the last agent's dataset. First,
        \begin{align*}
            \|\frac{\widetilde{X}^T\widetilde{y}}{n}\|_2
            &\leq\frac{1}{n}\|\widetilde{X}^{T}\|_2\|\widetilde{y}\|_2
            = \frac{1}{n}\sup_{v\in \mathcal{S}^{n-1}}\|\widetilde{X}^Tv\|_2\|\widetilde{y}\|_2\\
            &=\frac{1}{n}\sup_{v\in \mathcal{S}^{n-1}}\|\sum_{i\in [n]}\mathbf{\widetilde{x}}_iv_i\|_2\|\widetilde{y}\|_2
            \leq \frac{1}{n}\sup_{v\in \mathcal{S}^{n-1}}\sum_{i\in[n]}\|\mathbf{\widetilde{x}}_i\|_2|v_i|\|\widetilde{y}\|_2\\
            &\leq \frac{1}{n}\sup_{v\in \mathcal{S}^{n-1}}\sum_{i\in [n]}d^{\frac{1}{4}}\|\widetilde{\mathbf{x}}_i\|_4|v_i|\|\widetilde{y}\|_2
            \leq \frac{d^{\frac{1}{4}}\tau_1\tau_2}{\sqrt{n}}\sup_{v\in \mathcal{S}^{n-1}}\sum_{i\in [n]}|v_i|\leq d^{\frac{1}{4}}\tau_1\tau_2.
        \end{align*}
        Then we bound the sensitivity of $\|\frac{\widetilde{X}^T\widetilde{y}}{n}\|_2$,
        \begin{align*}
            \|\frac{\widetilde{X}^T\widetilde{y}}{n}-\frac{\widetilde{X}^{\prime^T}\widetilde{y}^{\prime}}{n}\|_2
            =\frac{1}{n}\|\mathbf{\widetilde{x}}_n\widetilde{y}_n-\mathbf{\widetilde{x}}_n^{\prime}\widetilde{y}_n^{\prime}\|_2
            \leq \frac{1}{n}(\|\mathbf{\widetilde{x}}_n\|_2|\widetilde{y}_n|+\|\mathbf{\widetilde{x}}_n^{\prime}\|_2|\widetilde{y}_n^{\prime}|)
            \leq \frac{2d^{\frac{1}{4}}\tau_1\tau_2}{n}.
        \end{align*}
        For any nonzero vector $w\in \bb{R}^d$,
        \begin{align*}
            &\quad\|\frac{\widetilde{X}^T\widetilde{X}}{n}w\|_2
            =\|\frac{\widetilde{X}^T\widetilde{X}}{n}w-\Sigma w+\Sigma w\|_2\\
            &\geq \|\Sigma w\|_2-\|(\frac{\widetilde{X}^T\widetilde{X}}{n}-\Sigma)w\|_2\geq (\kappa_{2}-\|\frac{\widetilde{X}^T\widetilde{X}}{n}-\Sigma\|_2)\|w\|_2.
            \numberthis\label{lem:sensitivity_linear1}
        \end{align*}
        By Lemma~\ref{lem:cov_est_heavytailed}, when $\tau_1=\Theta((n/\log n)^{1/4})$, with probability at least $1-dn^{-C_0}$ we have $\|\frac{\widetilde{X}^T\widetilde{X}}{n}-\Sigma\|_2\leq 2\sqrt{\frac{R d\log n}{n}}$. Thus when $n$ is sufficiently large such that $2\sqrt{\frac{ Rd\log n}{n}}\leq \frac{\kappa_{2}}{2}$, we have $\|\frac{\widetilde{X}^T\widetilde{X}}{n}-\Sigma\|_2\leq \frac{\kappa_{2}}{2}$. Combining this inequality and~\eqref{lem:sensitivity_linear1} delivers that $\|\frac{\widetilde{X}^T\widetilde{X}}{n}w\|_2\geq \frac{\kappa_{2}}{2}\|w\|_2$, which implies $\|(\frac{\widetilde{X}^T\widetilde{X}}{n})^{-1}\|_2\leq \frac{2}{\kappa_{2}}$. Thus,
        \begin{align*}
            \|(\frac{\widetilde{X}^{T}\widetilde{X}}{n})^{-1}-(\frac{\widetilde{X}^{\prime^T}\widetilde{X}^{\prime}}{n})^{-1}\|_2
            &\leq \|(\frac{\widetilde{X}^T\widetilde{X}}{n})^{-1}\|_2\|(\frac{\widetilde{X}^{\prime^T}\widetilde{X}^{\prime}}{n})^{-1}\|_2\|\frac{\widetilde{X}^T\widetilde{X}}{n}-\frac{\widetilde{X}^{\prime^T}\widetilde{X}^{\prime}}{n}\|_2\\
            &\leq \frac{4}{\kappa_{2}^2}
            (\|\frac{\widetilde{X}^T\widetilde{X}}{n}-\Sigma\|_2+\|\Sigma-\frac{\widetilde{X}^{\prime^T}\widetilde{X}^{\prime}}{n}\|_2)
            \leq \frac{8}{\kappa_{2}^2}\sqrt{\frac{Rd\log n}{n}}.
        \end{align*}
        By applying the inequality $\|AB-A^{\prime}B^{\prime}\|_2= \|AB-AB^{\prime}+AB^{\prime}-A^{\prime}B^{\prime}\|_2 \leq \|A\|_2\|B-B^{\prime}\|_2+\|A-A^{\prime}\|_2\|B^{\prime}\|_2$ and setting $\tau_2=\Theta((n/\log n)^{1/8})$ we have
        \begin{align*}
         \|\hat{\theta}(D)-\hat{\theta}(D^{\prime})\|_2
            &=\|(\frac{\widetilde{X}^T\widetilde{X}}{n})^{-1}\frac{\widetilde{X}^T\widetilde{y}}{n}-(\frac{\widetilde{X}^{\prime^T}\widetilde{X}^{\prime}}{n})^{-1}\frac{\widetilde{X}^{\prime^T}\widetilde{y}^{\prime}}{n}\|_2\\
            &\leq\frac{8d^{\frac{1}{4}}\tau_1\tau_2}{\kappa_{2}^2}\sqrt{\frac{Rd\log n}{n}}+\frac{4d^{\frac{1}{4}}\tau_1\tau_2}{\kappa_{2}n}
            = O(d^{\frac{3}{4}}(\frac{\log n}{n})^{\frac{1}{8}}).
            \numberthis\label{lem:sensitivity_linear2}
        \end{align*}
    \end{proof}
    
    \begin{proof} [{\bf Proof of Lemma~\ref{lem:accuracy_linear}}]
        \label{proof_lem_accuracy_linear}
        Note that by the proof of Lemma~\ref{lem:sensitivity_linear},
        when $n$ is sufficiently large such that $2\sqrt{\frac{Rd\log n}{n}}\leq \frac{\kappa_{2}}{2}$, with probability at least $1-dn^{-C_0}$, we have $\|(\frac{\widetilde{X}^T\widetilde{X}}{n})^{-1}\|_2\leq \frac{2}{\kappa_{2}}$. Thus,
        \begin{align*}
          \|\hat{\theta}(D)-\theta^{*}\|_2
            &=\|\theta^{*}-\big(\frac{\widetilde{X}^T\widetilde{X}}{n}\big)^{-1}\frac{\widetilde{X}^T\widetilde{y}}{n}\|_2 \\ &\leq\|\big(\frac{\widetilde{X}^T\widetilde{X}}{n}\big)^{-1}\|_{2} \|\big(\frac{\widetilde{X}^T\widetilde{X}}{n}\big)\theta^{*}-\frac{\widetilde{X}^T\widetilde{y}}{n}\|_2 \\
            & \leq \frac{2}{\kappa_2}
            \|\frac{\widetilde{X}^T}{n}\{\widetilde{X}\theta^{*}-\widetilde{y}\}\|_{2}\\
            & \leq\frac{2\sqrt{d}}{\kappa_{2}}\max_{j\in [d]}|\frac{1}{n}\sum_{i=1}^{n}\widetilde{x}_{ij}(\langle \widetilde{\mathbf{x}}_i,\theta^{*}\rangle-\widetilde{y}_i)|.
            \numberthis\label{lem:acc_linear1}
        \end{align*}
        Next we bound  $|\frac{1}{n}\sum_{i=1}^{n}\widetilde{x}_{ij}(\langle\widetilde{\mathbf{x}}_i,\theta^{*}\rangle-\widetilde{y}_i)|$.
        \begin{align*}
            & \quad |\frac{1}{n}\sum_{i=1}^{n}\widetilde{x}_{ij}(\langle\widetilde{\mathbf{x}}_i,\theta^{*}\rangle-\widetilde{y}_i)|\\
            &\leq|\frac{1}{n}\sum_{i=1}^n\widetilde{x}_{ij}\langle\widetilde{\mathbf{x}}_i,\theta^{*}\rangle-\bb{E}[\widetilde{x}_{ij}\langle\widetilde{\mathbf{x}}_i,\theta^{*}\rangle]|
            +|\bb{E}[\widetilde{x}_{ij}\langle\widetilde{\mathbf{x}}_i,\theta^{*}\rangle]-\bb{E}[\widetilde{x}_{ij}\widetilde{y}_i]|+|\frac{1}{n}\sum_{i=1}^{n}\widetilde{x}_{ij}\widetilde{y}_i-\bb{E}[\widetilde{x}_{ij}\widetilde{y}_i]|\\
            &\equiv \text{I}+\text{II}+\text{III}.
        \end{align*}
        Under the assumption that $\bb{E}(\nu^T\mathbf{x}_i)^4\leq R_1$ for any $\nu\in \mathcal{S}^{d-1}$, if  $\nu=\frac{\mathbf{x}_i}{\|\mathbf{x}_i\|_2}$, then $\bb{E}\|\mathbf{x}_i\|_2^4\leq R_1$; if $\nu=e_j$ for any $j\in [d]$ ($e_j$ is the unit vector whose $j$-th element is $1$), then $\bb{E}|x_{ij}|^4\leq R_1$ and thus $\bb{E}\|\mathbf{x}_i\|_4^4=\sum_{j=1}^{d}\bb{E}|x_{ij}|^4\leq dR_1$. Since
        \begin{align*}
            |\frac{1}{n}(\widetilde{x}_{ij}\langle\widetilde{\mathbf{x}}_i, \theta^*\rangle-\bb{E}[\widetilde{x}_{ij}\langle\widetilde{\mathbf{x}}_i, \theta^{*}\rangle])|
            \leq\frac{2}{n}|\widetilde{x}_{ij}|\|\widetilde{\mathbf{x}}_i\|_2\|\theta^{*}\|_2
            \leq\frac{2d^{\frac{1}{4}}\tau_1^2\tau_{\theta}}{n},
        \end{align*}
        and
        \begin{align*}
            \mathrm{var}(\frac{1}{n}(\widetilde{x}_{ij}\langle\widetilde{\mathbf{x}}_i,\theta^*\rangle-\bb{E}[\widetilde{x}_{ij}\langle\widetilde{\mathbf{x}}_i, \theta^{*}\rangle]))
            \leq\frac{1}{n^2}\bb{E}|\widetilde{x}_{ij}|^2\|\mathbf{\widetilde{x}}_i\|_2^2\|\theta^{*}\|_2^2
            \leq \frac{\tau_{\theta}^2}{n^2}\bb{E}\|\mathbf{x}_i\|_2^4\leq \frac{\tau_{\theta}^2R_1}{n^2},
        \end{align*}
        by Bernstein's inequality (Lemma~\ref{lem:Bernstein_ineq}), we obtain for any $t>0$,
        \begin{align*}
            \bb{P}\left(\text{I}>\frac{4d^{\frac{1}{4}}\tau_{\theta}\tau_1^2t}{3n}+\sqrt{\frac{2R_1\tau_{\theta}^2t}{n}}\right)\leq 2\exp(-t).
            \numberthis\label{lem:acc_linear2}
        \end{align*}
        Next we bound $\text{II}$.
        \begin{align*}
            \text{II}
            &=|\bb{E}\widetilde{x}_{ij}(\langle\widetilde{\mathbf{x}}_i,\theta^{*}\rangle-\widetilde{y}_i)|\\
            &\leq|\bb{E}\widetilde{x}_{ij}\langle\widetilde{\mathbf{x}}_i-\mathbf{x}_i,\theta^{*}\rangle|+|\bb{E}\widetilde{x}_{ij}(\langle\mathbf{x}_i,\theta^{*}\rangle-y_i)|+|\bb{E}\widetilde{x}_{ij}(y_i-\widetilde{y}_i)|\\
            &\leq \bb{E}|\widetilde{x}_{ij}|\|\mathbf{\widetilde{x}}_i-\mathbf{x}_i\|_2\|\theta^{*}\|_2+|\bb{E}_{\mathbf{x}_i}[\widetilde{x}_{ij}\bb{E}_{y_i}[\langle\mathbf{x}_i, \theta^{*}\rangle-y_i]]|+\bb{E}|\widetilde{x}_{ij}(y_i-\widetilde{y}_i)|\\
            &\leq \tau_{\theta}\bb{E}|\widetilde{x}_{ij}|\|\mathbf{\widetilde{x}}_i-\mathbf{x}_i\|_2\mathbf{1}_{\{\|\mathbf{x}_i\|_4>\tau_1\}}+
            \bb{E}|\widetilde{x}_{ij}(y_i-\widetilde{y}_i)\mathbf{1}_{\{|y_i|>\tau_2\}}|\\
            &\leq \tau_{\theta}\sqrt{\bb{E}|\widetilde{x}_{ij}|^2\|\widetilde{\mathbf{x}}_i-\mathbf{x}_i\|_2^2}\sqrt{\bb{P}(\|\mathbf{x}_i\|_4>\tau_1)}+\sqrt{\bb{E}|\widetilde{x}_{ij}|^2|y_i-\widetilde{y}_i|^2}\sqrt{\bb{P}(|y_i|>\tau_2)}\\
            &\leq \tau_{\theta}\sqrt{\bb{E}\|\widetilde{\mathbf{x}}_i\|_2^2\|\widetilde{\mathbf{x}}_i-\mathbf{x}_i\|_2^2}\sqrt{\frac{\bb{E}\|\mathbf{x}_i\|_4^4}{\tau_1^4}}+
            (\bb{E}|\widetilde{x}_{ij}|^4)^{\frac{1}{4}}(\bb{E}|y_i-\widetilde{y}_i|^4)^{\frac{1}{4}}\sqrt{\frac{\bb{E}|y_i|^4}{\tau_2^4}}\\
            &\leq \tau_{\theta}\sqrt{\bb{E}\|\mathbf{x}_i\|_2^4}\sqrt{\frac{\bb{E}\|\mathbf{x}_i\|_4^4}{\tau_1^4}}+(\bb{E}|x_{ij}|^4)^{\frac{1}{4}}(\bb{E}|y_i|^4)^{\frac{3}{4}}\frac{1}{\tau_2^2}\\
            &\leq\sqrt{d}\tau_{\theta}R_1\frac{1}{\tau_1^2}+R_1^{\frac{1}{4}}R_2^{\frac{3}{4}}\frac{1}{\tau_2^2}.
            \numberthis \label{lem:acc_linear3}
        \end{align*}
        Since $|\frac{1}{n}(\widetilde{x}_{ij}\widetilde{y}_i-\bb{E}[\widetilde{x}_{ij}\widetilde{y}_i])|\leq \frac{2}{n}\tau_1\tau_2, \mathrm{var}(\frac{1}{n}(\widetilde{x}_{ij}\widetilde{y}_i-\bb{E}[\widetilde{x}_{ij}\widetilde{y}_i]))\leq \frac{\sqrt{R_1R_2}}{n^2}$, by Bernstein's inequality (Lemma~\ref{lem:Bernstein_ineq}), we obtain for any $t>0$,
        \begin{align*}
            \bb{P}\left(\text{III}>\frac{4\tau_1\tau_2t}{3n}+\sqrt{\frac{2\sqrt{R_1R_2}t}{n}}\right)
            \leq 2\exp(-t). \numberthis\label{lem:acc_linear4}
        \end{align*}
        Let $\tau_1=\Theta((n/\log n)^{1/4})$, $\tau_2 =\Theta((n/\log n))^{1/8}$ and $t=\delta\log n$ for any $\delta>0$.  Combining~\eqref{lem:acc_linear2},~\eqref{lem:acc_linear3}, and~\eqref{lem:acc_linear4} delivers that for some constant $C=O(1)$ and for any $\delta>1$,
        \begin{align*}
            \bb{P}\left(\text{I}+\text{II}+\text{III}> Cd^{\frac{1}{2}}(\frac{\delta\log n}{n})^{\frac{1}{4}}\right)\leq 1-4n^{-\delta}.
        \end{align*}
        By the union bound for all $j\in [d]$ and~\eqref{lem:acc_linear1}, it holds that
        \begin{align*}
            \|\theta^{*}-\big(\frac{\widetilde{X}^T\widetilde{X}}{n}\big)^{-1}\frac{\widetilde{X}^T\widetilde{y}}{n}\|_2\leq Cd(\frac{\delta\log n}{n})^{\frac{1}{4}},
        \end{align*}
        with probability at least $1-4dn^{-\delta}-dn^{-C_0}$.
    \end{proof}

    \begin{proof}[{\bf Proof of Theorem~\ref{thm:privacy_linear}}]
        The proof is the same as the proof of Theorem~\ref{thm:privacy_glm}. We omit it here for simplicity.
    \end{proof}

    \begin{proof}[{\bf Proof of Theorem~\ref{thm:truthfulness_linear}}]
        Similar to the proof of Theorem~\ref{thm:truthfulness_glm}, it is not difficult to compute the maximum expected increased payment by misreporting to the analyst
        \begin{align*}
            &\quad \bb{E}[\pi_i(\hat{D}_i, \sigma(D^b, c^b))|D_i,c_i]-\bb{E}[\pi_i(D_i, \sigma(D^b, c^b))|D_i,c_i]\\
            &\leq a_2\|\widetilde{\mathbf{x}}_i\|_2^2
            \|\bb{E}[\bar{\theta}^P(\hat{D}^b)
            -\bb{E}_{\theta\sim p(\theta|D_i)}[\theta]|D_i,c_i]\|_2^2\\
            &\leq a_2\sqrt{d}\tau_1^2
            (
            \bb{E}\|\hat{\theta}(\hat{D}^b)-\hat{\theta}(D^b)\|_2
            +\bb{E}_{\theta\sim p(\theta|D_i), X^b}[\bb{E}_{y^b}[\|\hat{\theta}(X^b,y^b)-\theta\|_2|X^b, \theta]]
            +\bb{E}\|v_b\|_2
            )^2\\
            &\leq a_2\sqrt{\frac{dn}{\log n}}(\alpha n \Delta_{n/2}
            +Cd(\frac{\log n}{n})^{\frac{1}{4}}+
            \frac{d\Delta_{n/2}}{\varepsilon})^2.
        \end{align*}
        The decrease in privacy cost is bounded above by $\tau_{\alpha,\beta}F(2\varepsilon, \gamma_n+2\gamma_{n/2})$. Thus agent $i$'s total incentive to deviate is bounded above by
        \begin{align*}
            \eta=O(a_2\sqrt{\frac{dn}{\log n}}(\alpha n \Delta_{n/2}
            +Cd(\frac{\log n}{n})^{\frac{1}{4}}+
            \frac{d\Delta_{n/2}}{\varepsilon})^2
            +\tau_{\alpha,\beta}F(2\varepsilon, \gamma_n+2\gamma_{n/2})).
        \end{align*}
    \end{proof}
    
    \begin{proof}[{\bf Proof of Theorem~\ref{thm:accuracy_linear}}]
        Similar to the proof of Theorem~\ref{thm:accuracy_glm}, we have
        \begin{align*}
            \bb{E}\|\bar{\theta}^P(\hat{D})-\theta^{*}\|_2^2
            &\leq \bb{E}\|\hat{\theta}^P(\hat{D})-\theta^{*}\|_2^2\\
            &\leq 2\bb{E}\|\hat{\theta}^P(\hat{D})-\hat{\theta}(D)\|_2^2
            +2\bb{E}\|\hat{\theta}(D)-\theta^{*}\|_2^2\\
            &\leq 2((\alpha n\Delta_n)^2+d(d+1)(\frac{\Delta_{n}}{\varepsilon})^2
            +Cd^2(\frac{\log n}{n})^{\frac{1}{2}})\\
            &=\widetilde{O}(\alpha^2d^{\frac{3}{2}}n^{\frac{7}{4}}
            +d^{\frac{7}{2}}n^{-\frac{1}{4}}\varepsilon^{-2}
            +d^2n^{-\frac{1}{2}}).
        \end{align*}
    \end{proof}
    
    \begin{proof}[{\bf Proof of Theorem~\ref{thm:rationality_linear}}]
        Both of  two inputs of $B_{a_1, a_2}(\cdot, \cdot)$, {\em i.e.}, $\langle\widetilde{\mathbf{x}}_i, \bar{\theta}^P(\hat{D}^b)\rangle$ and $\langle\widetilde{\mathbf{x}}_i, \bb{E}_{\theta\sim p(\theta|\hat{D}_i)}[\theta]\rangle$, are bounded above by $d^{\frac{1}{4}}\tau_1\tau_{\theta}$.
        Thus agent $i$'s expected utility is non-negative as long as
        \begin{align*}
            a_1
            &\geq a_2(d^{\frac{1}{4}}\tau_1\tau_{\theta}+
            2(d^{\frac{1}{4}}\tau_1\tau_{\theta})(d^{\frac{1}{4}}\tau_1\tau_{\theta})+(d^{\frac{1}{4}}\tau_1\tau_{\theta})^2)+\tau_{\alpha,\beta}F(2\varepsilon, \gamma_n+2\gamma_{n/2})\\
            &=a_2(d^{\frac{1}{4}}\tau_1\tau_{\theta}+
            3d^{\frac{1}{2}}\tau_1^2\tau_{\theta}^2)+\tau_{\alpha,\beta}F(2\varepsilon, \gamma_n+2\gamma_{n/2}).
        \end{align*}
    \end{proof}
    
    \begin{proof}[{\bf Proof of Theorem~\ref{thm:budget_linear}}]
        \begin{align*}
            \mathcal{B}\leq n(a_1+a_2(d^{\frac{1}{4}}\tau_1\tau_{\theta}+(d^{\frac{1}{4}}\tau_1\tau_{\theta})^2)).
            \numberthis\label{thm:budget_linear1}
        \end{align*}
        Substitue  $a_1=a_2(d^{\frac{1}{4}}\tau_1\tau_{\theta}+3d^{\frac{1}{2}}\tau_1^2\tau_{\theta}^2)+\tau_{\alpha,\beta}F(2\varepsilon, \gamma_n+2\gamma_{n/2})$ into~\eqref{thm:budget_linear1} and let $\tau_1=\Theta((n/\log n)^{1/4})$, then
        \begin{align*}
            \mathcal{B}=\widetilde{O}(n(a_2\sqrt{dn}+\tau_{\alpha,\beta}F(2\varepsilon, \gamma_n+2\gamma_{n/2}))).
        \end{align*}
    \end{proof}
    
    \begin{proof}[{\bf Proof of Corollary~\ref{cor:heavy_linear}}]
         For any $\delta\in (\frac{1}{9}, \frac{1}{8})$ and $c>0$, we set $\tau_1=\Theta((n/\log n)^{1/4})$, $\tau_2=\Theta((n/\log n)^{1/8})$, $\varepsilon=n^{-\delta}$, $\alpha=\Theta(n^{-1+\delta})$, $\beta=\Theta(n^{-c})$, $a_2=n^{-\frac{1}{2}-9\delta}$, $a_1=a_2(d^{\frac{1}{4}}\tau_1\tau_{\theta}+3d^{\frac{1}{2}}\tau_1^2\tau_{\theta}^2)+\tau_{\alpha, \beta}F(2\varepsilon, \gamma_n+2\gamma_{n/2})$. Then, by Theorem~\ref{thm:accuracy_linear}, the private estimator is $\widetilde{O}(n^{-\frac{1}{4}+2\delta})$-accurate.
         By Lemma~\ref{lem:bound_on_threshold} and the assumption that 
         $F(\varepsilon, \gamma)=(1+\gamma)\varepsilon^{9}$, we have
         $\tau_{\alpha,\beta}F(2\varepsilon, \gamma_n+2\gamma_{n/2})=\widetilde{O}(n^{-9\delta})$. Note 
         that $a_2(\alpha^2 d^2n^{\frac{9}{4}}+d^4n^{\frac{1}{4}}\varepsilon^{-2})=\widetilde{O}(n^{-\frac{1}{4}-7\delta})$. For any $\delta\in (\frac{1}{9}, \frac{1}{8})$, it holds that $-\frac{1}{4}-7\delta<-9\delta$. Thus, by Theorem~\ref{thm:truthfulness_linear}, we have $\eta=\widetilde{O}(n^{-9\delta})$. By the choice of of $a_1$ and Theorem~\ref{thm:rationality_linear}, the mechanism is individual rational for least $1-O(n^{-1+\delta})$ fraction of agents. Finally, by Theorem~\ref{thm:budget_linear}, the total expect budget is $\mathcal{B}=\widetilde{O}(n^{-9\delta+1})$. 
    \end{proof}

\end{document}